%% file: elsarticle-template-num.tex
\documentclass[times]{elsarticle}

\usepackage{amssymb}
\usepackage{amsmath}

\usepackage{graphicx}
\usepackage{geometry}
\usepackage{tabularx}
\usepackage{makecell} 
\usepackage{array}
\usepackage{threeparttable}
\usepackage{booktabs}
\usepackage{tikz}
\usepackage{hyperref} 

\usetikzlibrary{arrows.meta,positioning,calc,matrix,fit}
\usepackage{xcolor}
\usepackage{subcaption}
\newcommand{\fullcirc}{\tikz[baseline=-0.5ex]{\fill[blue!70!black] (0,0) circle (0.8ex);}}
\newcommand{\halfcirc}{%
  \tikz[baseline=-0.5ex]{%
    \draw[blue!70!black, thick] (0,0) circle (0.8ex);
    \begin{scope}
      \clip (0,0) circle (0.8ex);
      \fill[blue!70!black] (-0.8ex,-0.8ex) rectangle (0,0.8ex);
    \end{scope}
  }
}
\newcommand{\emptycirc}{\tikz[baseline=-0.5ex]{\draw[blue!70!black, thick] (0,0) circle (0.8ex);}}

\usepackage{ragged2e}

\journal{Medical Image Analysis}

\begin{document}

\begin{frontmatter}



\title{Adaptation of Foundation Models for Medical Image Analysis: Strategies, Challenges, and Future Directions}


\author{ Karma Phuntsho, Abdullah, Kyungmi Lee, Ickjai Lee, Euijoon Ahn} 

\affiliation{organization={James Cook University},
            addressline={14-88 McGregor Rd}, 
            city={Smithfield},
            postcode={4878}, 
            state={QLD},
            country={Australia}}

\begin{abstract}
Foundation models (FMs) have emerged as a transformative paradigm in medical image analysis, offering the potential to provide generalizable, task-agnostic solutions across a wide range of clinical tasks and imaging modalities. Their capacity to learn transferable representations from large-scale data has the potential to address the limitations of conventional task-specific models. However, adaptation of FMs to real-world clinical practice remains constrained by key challenges, including domain shifts, limited availability of high-quality annotated data, substantial computational demands, and strict privacy requirements. This review presents a comprehensive assessment of strategies for adapting FMs to the specific demands of medical imaging. We examine approaches such as supervised fine-tuning, domain-specific pretraining, parameter-efficient fine-tuning, self-supervised learning, hybrid methods, and multimodal or cross-modal frameworks. For each, we evaluate reported performance gains, clinical applicability, and limitations, while identifying trade-offs and unresolved challenges that prior reviews have often overlooked. Beyond these established techniques, we also highlight emerging directions aimed at addressing current gaps. These include continual learning to enable dynamic deployment, federated and privacy-preserving approaches to safeguard sensitive data, hybrid self-supervised learning to enhance data efficiency, data-centric pipelines that combine synthetic generation with human-in-the-loop validation, and systematic benchmarking to assess robust generalization under real-world clinical variability. By outlining these strategies and associated research gaps, this review provides a roadmap for developing adaptive, trustworthy, and clinically integrated FMs capable of meeting the demands of real-world medical imaging.
\end{abstract}




\end{frontmatter}



\input{01_intro}
\input{02_background}

\input{03_evolution_of_fm}

\input{04_adaptations}

\input{05_applications}
\input{06_trends}
\input{07_conclusion}




\bibliographystyle{elsarticle-num}   
\biboptions{sort&compress}

\input{elsarticle-template-num.bbl}\end{document}

%% file: 01_intro.tex
\section{Introduction}\label{sec1:intro}
Medical image analysis (MIA) has become a cornerstone of modern clinical workflows, playing a critical role in the diagnosis, prognosis, treatment planning, and monitoring of a wide range of diseases. Yet, despite major advances in imaging technology, the interpretation of medical images remains a complex, labor-intensive, and error-prone process. For instance, radiologists may miss up to 30\% of lung nodules on chest X-rays \cite{ref2_radiologiest_missing_case}, while inter-observer variability in diagnosis continues to undermine diagnostic consistency and reliability across tasks such as breast cancer screening and tumor grading \cite{ref3_interobserver_variability}. Moreover, several
persistent challenges hamper progress in MIA, including the high cost of expert annotation, the scarcity of large, high-quality labeled datasets, and stringent patient privacy requirements \cite{ref9_fm_medical_comprehensive_survey, ref12_challenges_per_fm}. These limitations emphasize the need for robust, automated, and generalizable solutions, an area where artificial intelligence, particularly deep learning has demonstrated considerable promise. For instance, convolutional neural networks (CNNs) \cite{ref4_cnn} and U-Net architectures \cite{ref5_unet,ref6_unetr,ref7_swinunetr} have long been considered the backbone of MIA, demonstrating strong performance in various medical tasks. CNNs automatically learn hierarchical spatial features from raw pixel data using convolutional filters, while U-Nets employ a distinctive encoder-decoder structure with skip connections that preserve spatial resolution, making them particularly effective for dense segmentation. Although these models have enhanced diagnostic accuracy and efficiency, their broader clinical adoption remains constrained by the scarcity of annotated data, poor generalization under domain shifts, task-specific design, high computational requirements, and limited interpretability \cite{ref9_fm_medical_comprehensive_survey}. 

In contrast, \textit{foundation models} (FMs) \cite{ref10_fm} have emerged as a transformative paradigm, offering greater adaptability, scalability, and cross-domain generalization. Their impact is most visible in natural language processing (NLP), where models such as GPT-4 \cite{ref16_gpt4}, LLaMA \cite{ref15_llama}, and more recently GPT-5 \cite{ref161_2025gpt5}  have shown unprecedented generalization and adaptability across a wide spectrum of tasks. These models excel in zero-shot learning (solving new tasks without prior examples), few-shot learning (adapting from only a handful of labeled examples), and instruction-following scenarios \cite{ref18_overview_of_llm}, achieving state-of-the-art (SOTA) performance across diverse applications. Their success has been driven by advances in computational hardware, availability of large training datasets, and the introduction of transformer architecture \cite{ref10_fm}. Introduced by Vaswani et al. \cite{ref11_transformer}, transformers model long-range dependencies through self-attention mechanisms, offering greater flexibility in capturing complex relationships across input elements. This architectural breakthrough has led to the development of models such as vision transformer (ViT) \cite{ref19_vit}, swin transformer \cite{ref20_swin}, segment anything model (SAM) \cite{ref21_sam}, as well as multimodal models like contrastive language image pre-training (CLIP) \cite{ref22_clip} and bootstrapping language-image pre-training (BLIP) \cite{ref23_blip}. These models learn robust and generalizable visual or multimodal representations, enabling effective transfer to a wide variety of downstream tasks. Their ability to perform zero- and few-shot adaptation with minimal labeled data makes them particularly valuable in MIA, where annotated datasets remain scarce and costly to obtain.


The rapid progress of FMs in NLP and computer vision raises an important question: how can these powerful and general-purpose models be effectively adapted to the unique and demanding field of MIA? Despite the remarkable capabilities of FMs, medical practice still relies heavily on task-specific models for applications such as disease diagnosis and planning. This reliance is primarily due to the distinctive challenges presented by the diverse medical imaging modalities, which differ fundamentally from natural images in terms of structure, semantics, and acquisition characteristics. Moreover, medical datasets tend to be small, multi-dimensional, highly imbalanced, and expensive to annotate due to the need for specific domain expertise \cite{ref1_3d_ssl_methods_medical}. These factors, combined with the high-stakes nature of clinical decision-making, mean that direct application of generalist models is insufficient and may even introduce risk \cite{ref12_challenges_per_fm}. Yet, the core strengths of FMs present an unprecedented opportunity to bridge the gap between the general-purpose models and the specialized requirements of MIA through targeted adaptations. Recent studies have shown that FMs pretrained with self-supervised learning (SSL) can be fine-tuned with minimal labeled data to achieve strong performance on medical tasks \cite{ref25_ssl_swin_trans, ref26_ssl_mae_mia}. SSL approaches enable models to learn useful representations directly from unlabeled data and form the backbone of multimodal models such as CLIP \cite{ref22_clip} and BioCLIP \cite{ref23_blip}, supporting advanced applications such as automated radiology report generation and cross-modal image-text retrieval \cite{ref29_multimodal_report_generation, ref30_contrastive_learning}. Moreover, domain-adapted FMs such as MedSAM \cite{ref31_medsam} and MedCLIP \cite{ref32_medclip} further demonstrate how fine-tuning on medical datasets, or introducing architectural modifications can transform generalist models into specialized tools for healthcare.

However, directly adapting FMs pretrained on natural images to the medical domain is not straightforward, as the mismatch in data characteristics often results in suboptimal performance \cite{ref31_medsam}. The unique characteristics of medical data and the distinct requirements of imaging tasks, combined with the clinical imperative for accuracy and reliability, necessitate careful adaptation and rigorous validation. Motivated by these challenges, this review examines how FMs can be effectively and responsibly tailored for MIA. Our objectives are threefold: (1) to critically assess current adaptation strategies and their reported clinical performance; (2) to identify key methodological and clinical gaps between general-purpose vision models and the unique requirements of medical imaging; and (3) to evaluate the effectiveness of adaptation techniques in bridging these gaps across core imaging tasks. Ultimately, by advancing both theoretical understanding and practical methodologies, we aim to support the development of FMs that can genuinely enhance MIA and improve patient care.

While the literature on FMs in MIA is rapidly expanding, existing reviews often remain broad in scope, focusing primarily on architectures, generic challenges, and high-level applications, while overlooking the concrete mechanisms of adaptation.
For example, He et al. \cite{ref35_fm_advancing_healthcare} provided a panoramic methodological overview of healthcare FMs, spanning pre-training, adaptation, datasets, applications, and future challenges, but their focus remains broad, prioritizing taxonomy and future outlook over task-specific depth. Similarly, Khan et al. \cite{ref33_comprehensive_survey_fm} presented a broad survey of FMs across medicine, offering a taxonomy spanning NLP, vision, omics, graphs, and multimodal applications, but their analysis remains largely descriptive, highlighting flagship models and domain-wide challenges without examining adaptation workflows in detail.
In contrast, our review centers adaptation strategies and critically examines their clinical applicability across core tasks such as classification, segmentation, and detection as structured in Table \ref{tab:review_comparison}. By moving beyond high-level overviews, we provide a task-specific analysis of supervised fine-tuning, parameter-efficient fine-tuning (PEFT), SSL, hybrid approaches (e.g., hybrid PEFT, hybrid SSL, and hybrid architectures), and multimodal or cross-modal adaptations. Prior reviews \cite{ref36_vfm_mia,ref37_fm_misegmentation,ref12_challenges_per_fm, ref34_generalist_fm} have largely focused on downstream applications or broad challenges, and often neglected these recent advances, which are increasingly central to efficient and scalable adaptation. To address these gaps, our review makes the following contributions. First, we place adaptation strategies at the center of the analysis, moving beyond broad architectural overviews to critically examine how FMs can be tailored to the unique demands of medical imaging. Specifically:
\begin{enumerate}
    \item We critically examine and compare major adaptation strategies including supervised fine-tuning, PEFT, SSL, emerging hybrid strategies, and multimodal/cross-modal approaches.
    \item  For each strategy, we evaluate applications, reported performance gains, clinical suitability, and limitations, highlighting practical trade-offs and unresolved challenges that are often overlooked in prior reviews.
    \item We provide targeted insights into how these adaptation methods perform across classification, segmentation, and detection tasks, linking strategies to clinical relevance and demonstrating their role in bridging methodological and application-level gaps.
    \item We deliver clear, actionable recommendations for researchers and practitioners on selecting and adapting FMs responsibly and efficiently, emphasizing both methodological rigor and clinical robustness to accelerate trustworthy adoption in clinical practice.
    
\end{enumerate}
\begin{table}[!htbp]
\centering
\caption{Comparative summary of recent survey and review articles on FMs in medical imaging, indicating their coverage of key adaptation strategies including supervised fine-tuning, PEFT, self-supervised learning, and multimodal/cross-modal methods, as well as core medical imaging applications.}
\renewcommand{\arraystretch}{1.1}
\setlength{\tabcolsep}{3pt}
\begingroup
\fontsize{9}{11}\selectfont
\begin{tabularx}{\textwidth}{%
    >{\raggedright\arraybackslash}X
    >{\centering\arraybackslash}p{2cm}
    >{\centering\arraybackslash}p{1.5cm}
    >{\centering\arraybackslash}p{2cm}
    >{\centering\arraybackslash}p{2.3cm}
    >{\centering\arraybackslash}p{2cm}
}
    \toprule
    \textbf{Survey Reference} & \textbf{Supervised Fine-tuning} & \textbf{PEFT} & \textbf{Self-Supervised} & \textbf{Multimodal/\newline Cross-Modal} & \textbf{Core \newline Applications} \\
    \midrule
    Khan et al., 2025 \cite{ref33_comprehensive_survey_fm} & \halfcirc & \emptycirc & \halfcirc & \halfcirc & \fullcirc \\
    He et al., 2025 \cite{ref35_fm_advancing_healthcare}   & \halfcirc & \halfcirc & \halfcirc & \halfcirc & \fullcirc \\
    Liang et al., 2025 \cite{ref36_vfm_mia}                & \halfcirc & \halfcirc & \emptycirc & \fullcirc & \fullcirc \\
    Lee et al., 2024 \cite{ref37_fm_misegmentation}        & \halfcirc & \halfcirc & \emptycirc & \emptycirc & \fullcirc \\
    Zhang and Metaxas, 2024 \cite{ref12_challenges_per_fm} & \emptycirc & \emptycirc & \emptycirc & \emptycirc & \fullcirc \\
    Moor et al., 2023 \cite{ref34_generalist_fm}           & \emptycirc & \emptycirc & \halfcirc & \halfcirc & \fullcirc \\
    \textbf{Ours}                                          & \textbf{\fullcirc} & \textbf{\fullcirc} & \textbf{\fullcirc} & \textbf{\fullcirc} & \textbf{\fullcirc} \\
    \bottomrule
\end{tabularx}

\vspace{0.5ex}
\noindent\textit{Note:}~\fullcirc~denotes all aspects are covered, \halfcirc~represents partial coverage, and \emptycirc~indicates the aspect is not discussed.
\endgroup
\label{tab:review_comparison}
\end{table}

The remaining section of this review is structured as follows: Section \ref{sec2:background} establishes the context by introducing FMs in the scope of MIA and outlining unique characteristics of medical imaging modalities, as well as introducing the evaluation metrics relevant to core tasks addressed in this review. Section \ref{sec3:evolution} traces the evolution of FMs, highlighting key architectural advances and pretraining paradigms.  Section \ref{sec4:adaptations} forms the heart of our review, presenting adaptation techniques, from supervised fine-tuning to parameter-efficient methods, self-supervised paradigms to multimodal and 3D extensions, and critically assessing their strengths, weaknesses, and practical considerations. Section \ref{sec5:applications} explores applications of FMs across core medical imaging tasks, followed by Section \ref{sec6:trends}, which discusses emerging trends and future directions in the field. Finally, Section \ref{sec7:conclusion} concludes the review with a synthesis of key insights.

%% file: 02_background.tex
\section{Background and Preliminaries} \label{sec2:background}
This section provides the foundational context for understanding the adaptation of foundation models in MIA. It introduces FMs and their role in MIA, reviews major imaging modalities, outlines core analysis tasks, and details key performance metrics that inform model evaluation and adaptation.

\subsection{Foundation Models for Medical Image Analysis}
The term FM was first introduced by Bommasani et al. \cite{ref10_fm} to describe large-scale neural networks pretrained on massive and diverse datasets, enabling them to learn highly generalizable representations. Unlike conventional models that are typically trained from scratch or fine-tuned on narrow, task-specific datasets, FMs are pretrained on massive and diverse datasets using self-supervised and, in many cases, multimodal objectives. This enables them to learn rich and transferable representations that can support generalization across tasks and domains. Furthermore, multimodal FMs have demonstrated the ability to integrate medical images, clinical text, and other patient data, enhancing contextual understanding and interpretability \cite{ref38_emergent_llm}. A defining feature of FMs is their emergent behavior, whereby they demonstrate abilities and generalization far beyond those explicitly present in their training data \cite{ref10_fm}. For example, models such as GPT-4 in NLP and SAM in computer vision have shown remarkable zero-shot and few-shot performance, challenging conventional boundaries between pretraining and task-specific fine-tuning \cite{ref16_gpt4,ref21_sam}. The evolution of multimodal FMs further highlights the flexibility and potential to unify different modalities under a single architecture.

However, the adaptation of FMs also poses several critical challenges. Their sheer scale and adaptability raise important concerns regarding transparency, fairness, and computational cost, particularly in high-stakes domain like medical imaging \cite{ref9_fm_medical_comprehensive_survey, ref37_fm_misegmentation}. Moreover, adapting FMs to the medical field is uniquely challenging due to the diversity and complexity of medical imaging modalities (see subsection \ref{modalities}), which differ substantially from natural images not only in semantics but also in underlying physical properties. Medical images are often grayscale, low-contrast, noisy, volumetric, and require fine-grained precision, all factors that complicate the direct transfer of models pretrained on natural images \cite{ref12_challenges_per_fm}. For example, while SAM has demonstrated impressive performance in promptable segmentation and zero-shot generalization on natural images, its out-of-the-box performance on challenging medical tasks such as pancreas or spine segmentation remains suboptimal \cite{ref39_sam_for_mia, ref_40_sammed2d}. However, when adapted through fine-tuning with high-quality expert-annotated medical datasets \cite{ref31_medsam, ref157_medsam2,ref119_sammed3d}, augmented with architectural adapters \cite{ref53_medsa,ref127_masam,ref54_SAMed,ref146_mediclip}, or enhanced via targeted adaptation strategies \cite{ref147_promptmrg, ref55_autosam}, SAM has demonstrated significantly improved performance in medical imaging applications. These findings underscore both the promise and the fragility of transferring FMs to the medical domain, emphasizing the need for careful adaptation and rigorous validation.

\subsection{Overview of Medical Imaging Modalities} \label{modalities}
Medical imaging encompasses a diverse and rapidly evolving set of modalities, each shaped by distinct physical principles and designed for specific clinical applications. X-ray imaging (radiography) is among the most widely used techniques. It uses X-ray radiation to generate 2D projection images of internal structures, which are essential for assessing bone fractures and chest pathologies. Computed Tomography (CT) uses X-ray to generate high-resolution cross-sectional images, making it ideal for detecting lung nodules, vascular abnormalities, and skeletal injuries \cite{ref41_sam_for_miseg}. Magnetic Resonance Imaging (MRI) leverages magnetic fields and radio waves to produce high-contrast soft tissue images, making it particularly effective in detecting tumors and other abnormalities in the brain, spine, and other organs. Ultrasound (US) uses high-frequency sound waves to visualize soft tissues and is distinguished by its real-time imaging capability and excellent safety mechanism. It is particularly useful for prenatal and abdominal assessments. Positron Emission Tomography (PET) captures metabolic activity through radiotracers and is often combined with CT in hybrid imaging for oncology applications \cite{ref42_medical_modalities}. A full comparison of common modalities is shown in Table \ref{tab:image_modality}.

Despite their clinical value, these imaging modalities present distinct challenges for MIA and the adaptation of AI and FMs. The data generated by these systems vary in dimensionality from 2D to 3D (and sometimes 4D), from grayscale to multi-channel, and from static to dynamic, necessitating models that can handle differing spatial resolutions, contrast levels, and noise characteristics \cite{ref33_comprehensive_survey_fm}. Another critical characteristic of medical imaging datasets is the severe class imbalance, where only a small fraction of scans contain positive findings compared to a large volume of negative cases. This imbalance poses challenges for training robust models, as it can bias predictions toward normal cases unless explicitly addressed. In addition to these data-level issues, each modality introduces its own challenges. For instance, while MRI provides rich anatomical detail, it is susceptible to motion artifacts and exhibits high inter-scanner and inter-protocol variability. CT offers greater consistency but introduces radiation-induced noise and bias. Moreover, modalities such as US present unique challenges for automation as it is highly operator-dependent, resulting in inconsistent image quality and difficulties in standardizing datasets for model training and validation \cite{ref12_challenges_per_fm}. Consequently, adapting FMs for MIA requires strategies that address both dataset heterogeneity and imbalance as well as modality-specific artifacts, while still enabling models to generalize effectively across protocols, scanners, and clinical tasks.

\begin{table}[!htbp]
\centering
\caption{Summary of common medical imaging modalities, their primary clinical applications, and challenges for AI/FM adaptation.}
\renewcommand{\arraystretch}{1.12}
\setlength{\tabcolsep}{3pt}
\begingroup
\fontsize{9}{11}\selectfont 
\begin{tabularx}{\textwidth}{%
  >{\raggedright\arraybackslash}p{2.0cm} 
  >{\raggedright\arraybackslash}p{2.0cm} 
  >{\raggedright\arraybackslash}X
  >{\raggedright\arraybackslash}X
  >{\raggedright\arraybackslash}X
}
\toprule
\textbf{Modality} & \textbf{Dimensions} & \textbf{Contrast / Detail} & \textbf{Common Clinical Applications} & \textbf{Challenges for AI/FMs} \\
\midrule
\textbf{X-ray}      & 2D (projection)   & Good bone detail, limited soft tissue & Bone fractures, chest pathologies & Low soft-tissue contrast, overlapping structures \\
\textbf{CT}         & 3D (cross-sectional) & High spatial resolution, moderate soft-tissue contrast & Lung nodules, vascular disease, skeletal injuries & Radiation exposure, noise, scanner-related bias \\
\textbf{MRI}        & 2D/3D/4D (dynamic) & Excellent soft-tissue contrast, high anatomical detail & Tumors, brain/spine imaging, cardiac assessments & Motion artifacts, inter-scanner variability, long acquisition time \\
\textbf{Ultrasound} & 2D/3D (real-time) & Real-time soft-tissue imaging & Prenatal, abdominal, cardiac assessments & Operator dependence, variable quality, standardization challenges \\
\textbf{PET}        & 3D (functional)   & Metabolic activity, often combined with CT & Oncology, functional imaging & Low resolution, noisy, radiotracers required, costly \\
\bottomrule
\end{tabularx}
\endgroup
\label{tab:image_modality}
\end{table}

\subsection{Core Tasks in Medical Image Analysis}
MIA encompasses several fundamental tasks, each designed to extract clinically relevant information from complex imaging data. Within the clinical workflow, these tasks often form a sequential pipeline, beginning with preprocessing, followed by detection and localization, segmentation, classification, and ultimately report generation to support clinical decision-making.

\textbf{Image Acquisition and Preprocessing:} Clinical imaging typically begins with the acquisition of image data which often contains noise, bias, and other artifacts specific to the modality and hardware used. Preprocessing is therefore essential to transform raw acquisition into standardized and reliable inputs for downstream analysis. While decades of methodological advances have improved preprocessing pipelines, variability across institutions, scanners, and patient populations remains a critical bottleneck, limiting the robustness of even the SOTA AI models \cite{ref33_comprehensive_survey_fm}.

\textbf{Detection and Localization:} The next step involves identifying clinically relevant regions or structures, such as lung nodules, lesions, or organs. While detection determines the presence and approximate position, localization extends this further by pinpointing the exact coordinates or spatial boundaries of the target \cite{ref43_survey_dl_mia}. This step is often high-stakes in clinical decision-making, as missed detection can lead to delayed diagnosis, while false positives (incorrectly identifying healthy tissue as disease) can trigger unnecessary interventions and anxiety. Although deep learning has advanced detection performance, it is often challenged by rare or subtle findings and severe class imbalance \cite{ref43_survey_dl_mia}. FMs with their capacity for large-scale pretraining and generalization could therefore offer a viable alternative if carefully adapted to the unique characteristics of medical imaging data.

\textbf{Segmentation:} Segmentation in medical imaging refers to the process of isolating and precisely delineating structures of interest such as organs or tumors. Traditionally, manual segmentation has been the gold standard, but it is labor-intensive and expensive, making it impractical for large-scale clinical use. While deep learning models have demonstrated remarkable potential in automating segmentation, their highly task-specific nature often limits their generalizability. As a result, performance can degrade significantly when these models are applied to new tasks, unseen disease types, or different imaging modalities, highlighting the ongoing challenge of achieving robust and reliable segmentation in real-world clinical practice \cite{ref31_medsam, ref_40_sammed2d}. Recently, domain-adapted FMs such as MedSAM \cite{ref31_medsam} or MedCLIP have demonstrated greater versatility, suggesting a path toward more robust and scalable solutions for clinical segmentation tasks.

\textbf{Classification:} Classification in medical imaging involves assigning clinical meaning, such as distinguishing benign and malignant lesions, grading tumor aggressiveness, or identifying disease subtypes \cite{ref43_survey_dl_mia}. These models form the core of computer-aided diagnosis (CAD) systems, providing diagnostic or prognostic labels for entire images or localized regions of interest. FMs, through improved representation learning and adaptability, hold potential to enhance classification across diverse modalities and tasks. However, their clinical reliability for nuanced or rare scenarios remains insufficiently validated, underscoring the need for rigorous testing in heterogeneous, real-world settings \cite{ref34_generalist_fm}.

\textbf{Report Generation and Decision Support:} The ultimate goal of MIA extends beyond detecting or classifying abnormalities to delivering actionable insights for clinical decision-making. Traditional reporting is labor-intensive and time-consuming, motivating interest in automated or semi-automated approaches that translate algorithmic outputs into structured summaries or narrative reports \cite{ref44_medical_report_generation}. In addition, recent advances in vision-language models and multimodal FMs have also enabled the development of systems capable of generating radiology reports directly from medical images \cite{ref43_survey_dl_mia}. While these systems could reduce radiologist workload and streamline documentation, real-world deployment remains limited due to ongoing challenges with interpretability, clinical validation, and the requirement for reports to meet professional practice standards \cite{ref44_medical_report_generation}.

\subsection{Key Performance Metrics }
Evaluating the performance of models requires a diverse set of metrics that capture predictive accuracy, robustness and clinical relevance. The choice of metric is critical for enabling fair comparisons across methods, assessing generalizability to real-world settings, and ensuring reproducibility across studies. Table~\ref{tab:metrics} summarizes commonly used metrics with their key characteristics and clinical relevance.

In classification, common metrics include accuracy, area under the receiver operating characteristic curve (AUC-ROC), precision, recall (sensitivity), and F1-score. Accuracy provides an overall summary of correct predictions among all cases but can be misleading in imbalanced datasets, where high accuracy may result from simply predicting the majority class \cite{ref45_mcc}. The AUC-ROC provides a more nuanced view by quantifying the trade-off between sensitivity (true positive rate) and specificity (true negative rate) across various decision thresholds \cite{ref46_auc_over_accuracy}. This makes it particularly valuable in medical settings, where the consequences of false positives and false negatives can differ significantly. Precision measures the proportion of positive predictions that are actually correct, while recall measures the proportion of actual positives correctly identified by the model. The F1-score is the harmonic mean of precision and recall, and is increasingly favored in recent literature for its balanced assessment of performance in imbalanced datasets, capturing both the ability to detect positive cases and avoid false alarms \cite{ref45_mcc}.

Segmentation quality is typically evaluated using metrics that assess regional overlap and boundary accuracy. Dice Similarity Coefficient (DSC) measures the overlap between the predicted segmentation and the ground truth, making it a widely adopted metric for assessing overall segmentation quality and reproducibility. However, DSC may be insensitive to boundary discrepancies and can overlook clinically relevant errors if boundaries are misaligned. Jaccard Index (Intersection over Union, IoU) similarly quantifies the proportion of shared area between the predicted and ground-truth regions, but imposes a stricter penalty for partial overlap, making it useful for multiclass segmentation tasks. In contrast Hausdorff Distance (HD) focuses on the maximum distance between the boundaries of the predicted and true segmentation. This metric is particularly sensitive to boundary errors and outliers, providing a stringent assessment of how well predicted boundaries align with the ground truth \cite{ref47_metrics}.

For detection tasks involving multiple object types, Average Precision (AP) and mean Average Precision (mAP) are widely used. AP measures the area under the precision-recall curve for each class, while mAP averages this value across all classes. These metrics provide a robust and threshold-independent assessment of detection performance, making them popular in both computer vision and medical imaging tasks \cite{ref48_detection_metrics}.
Ultimately, the choice of performance metrics should be guided by the clinical context, data characteristics, and the specific requirements of each task, as no single metric captures all aspects of model performance in medical imaging.

\begin{table}[t]
\centering
\begin{threeparttable}
\caption{Comparison of common evaluation metrics in MIA, their associated tasks, typical imaging modalities, key characteristics, and clinical relevance.}
\label{tab:metrics}
\renewcommand{\arraystretch}{1.1}
\setlength{\tabcolsep}{1.5pt}
\begingroup
\fontsize{9}{11}\selectfont
\begin{tabularx}{\textwidth}{%
  >{\raggedright\arraybackslash}p{2.0cm}  
  >{\raggedright\arraybackslash}p{2.0cm}  
  >{\raggedright\arraybackslash}p{2.0cm}  
  >{\raggedright\arraybackslash}X         
  >{\raggedright\arraybackslash}X         
}
\toprule
\textbf{Metric} & \textbf{Common Task(s)} & \textbf{Typical Modalities} & \textbf{Key Characteristics} & \textbf{Clinical Relevance} \\
\midrule
Accuracy & Classification & X-ray, CT, MRI &
Measures overall correctness, unreliable in imbalanced datasets &
Suitable for balanced cohorts, inappropriate for rare diseases \\
\addlinespace[2pt]
AUC-ROC & Classification & MRI, PET, CT &
Threshold-independent, evaluates sensitivity–specificity trade-off &
Ideal when false positives and false negatives have differing costs \\
\addlinespace[2pt]
Precision & Classification & CT, MRI, PET &
Penalizes false positives, focuses on positive prediction accuracy &
Minimizes unnecessary interventions or treatments \\
\addlinespace[2pt]
Recall (Sensitivity) & Classification & MRI, X-ray, PET &
Penalizes false negatives, prioritizes identifying true positives &
Critical for screening tasks, e.g., cancer detection \\
\addlinespace[2pt]
Specificity & Classification & CT, MRI, PET &
True-negative rate, complements recall &
Key when false positives risk harmful interventions \\
\addlinespace[2pt]
F1-score & Classification & MRI, CT, X-ray &
Balances precision and recall, robust to class imbalance &
Effective single metric for rare positive cases \\
\addlinespace[2pt]
Dice (DSC) & Segmentation & MRI, CT &
Overlap-based, volume-sensitive, may miss boundary errors &
Standard for assessing tumor or organ segmentation consistency \\
\addlinespace[2pt]
Jaccard (IoU) & Segmentation & MRI, CT, US &
Stricter than Dice, penalizes partial mismatches heavily &
Valuable for multiclass segmentation, e.g., multiple tissue types \\
\addlinespace[2pt]
Hausdorff Distance (HD) & Segmentation & MRI, CT &
Worst-case boundary discrepancy, sensitive to outliers &
Important for surgical or radiotherapy planning \\
\addlinespace[2pt]
Average Precision (AP) & Detection & X-ray, CT, US &
Threshold-independent, robust to imbalance &
Captures lesion-level detection quality \\
\addlinespace[2pt]
Mean AP (mAP) & Detection & X-ray, CT, US &
Aggregates detection across multiple categories &
Standard for multi-lesion or multi-disease detection \\
\bottomrule
\end{tabularx}
\begin{tablenotes}
\footnotesize
\item Note: Modalities are listed as those most commonly reported for the corresponding task, they are not intrinsic to the metric itself.
\end{tablenotes}
\endgroup
\end{threeparttable}
\end{table}

%% file: 03_evolution_of_fm.tex
\section{Evolution of Foundation Models} \label{sec3:evolution}
This section discusses advances in model architectures and pretraining paradigms, highlighting how these developments have shaped the evolution of FMs and their relevance to MIA.
\subsection{Advances in Model Architectures}
The evolution of FMs reflects a continuous shift in how data are represented and processed across both visual and multimodal domains. From early convolution-based designs focused on local spatial hierarchies, to transformer-driven architectures that capture long-range dependencies, and more recently to vision–language models that align visual and textual representations, each stage has redefined the balance between local feature extraction, global context modeling, and scalability across diverse applications.

\begin{figure}[!t]
    \centering
    \begin{subfigure}[b]{0.48\textwidth}
        \centering
        \includegraphics[width=\linewidth]{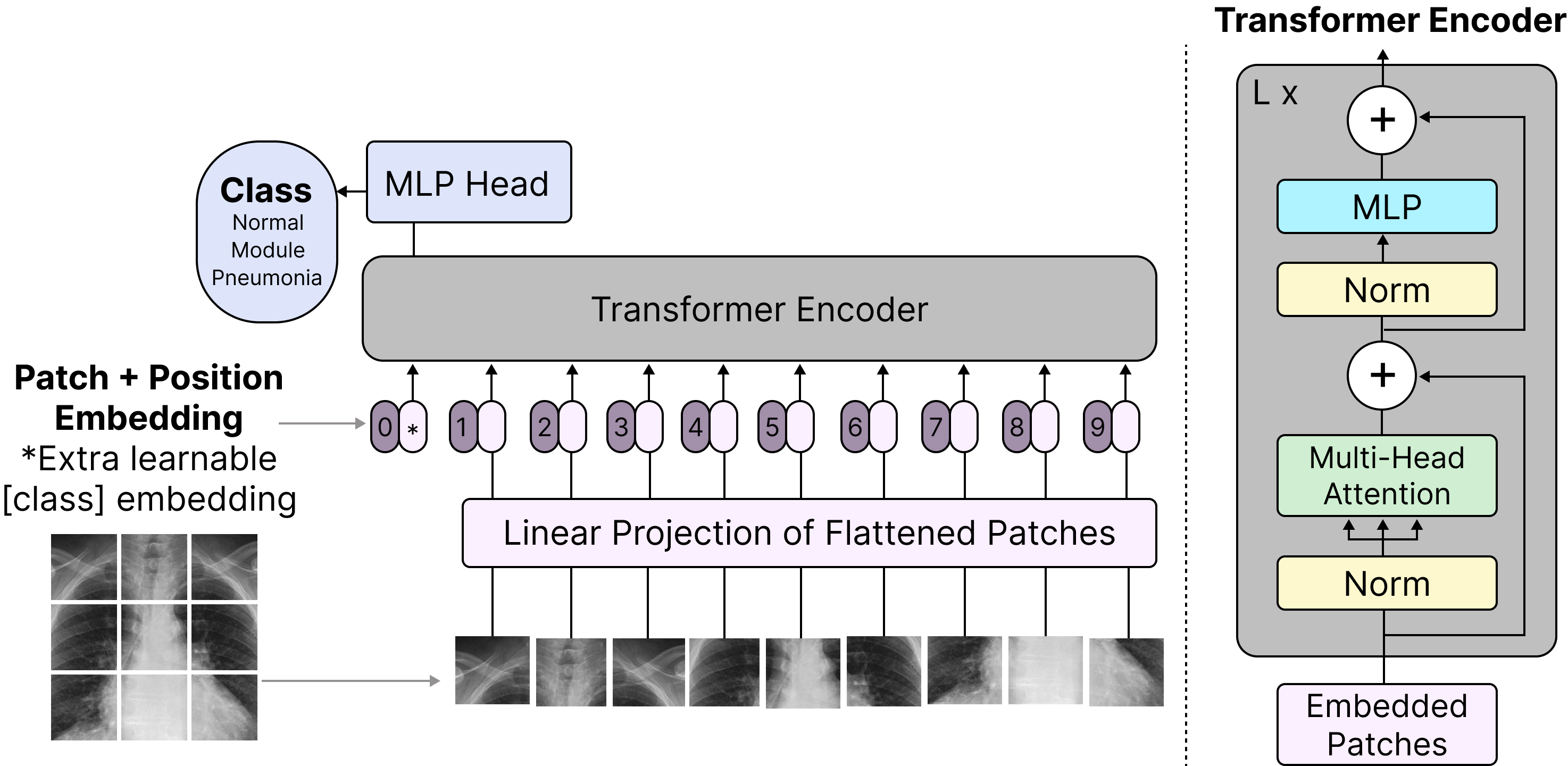}
        \caption{ViT}
        \label{fig:vit}
    \end{subfigure}
    \hfill
    \begin{subfigure}[b]{0.48\textwidth}
        \centering
        \includegraphics[width=\linewidth]{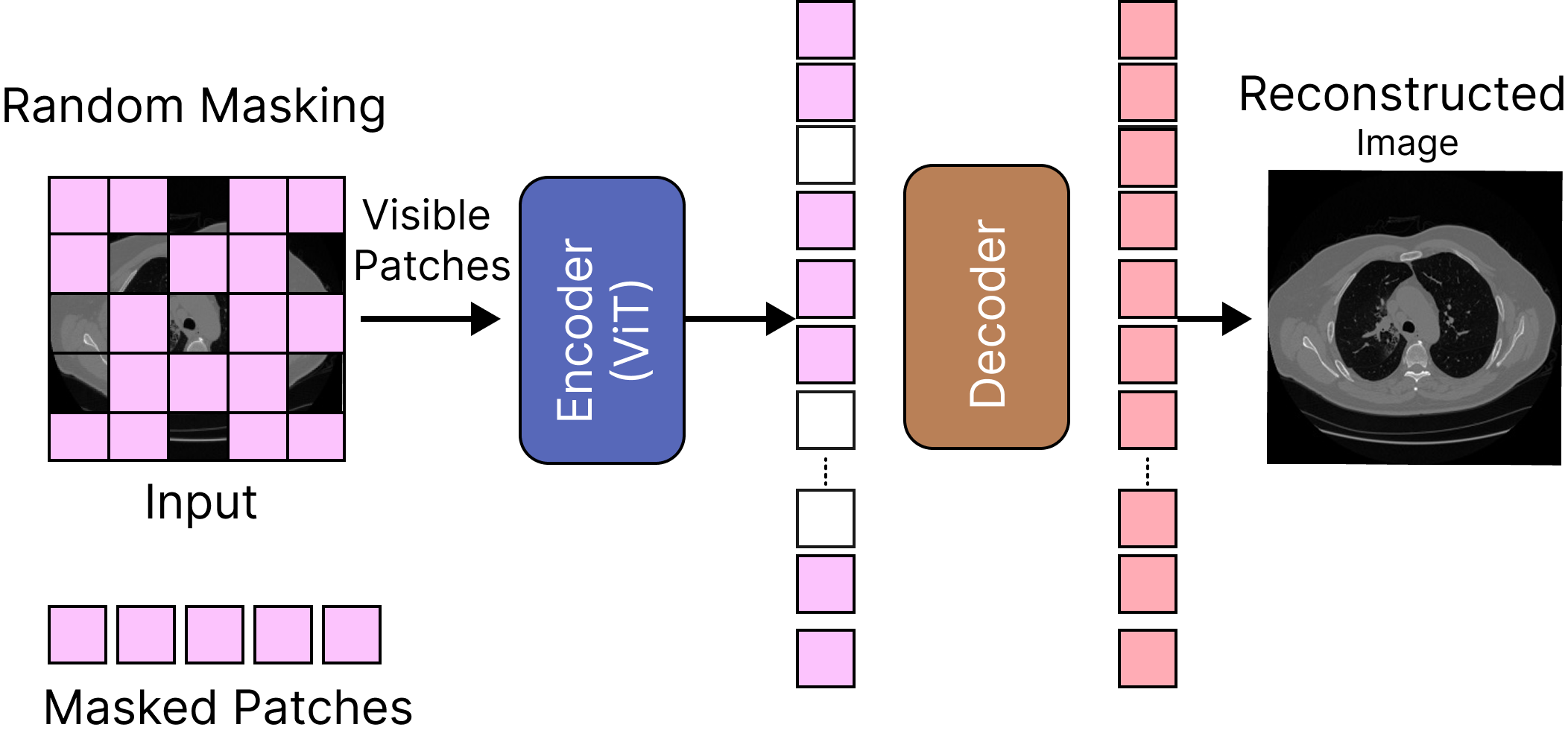}
        \caption{MAE}
        \label{fig:mae}
    \end{subfigure}

    \vspace{1em}
    \begin{subfigure}[b]{0.48\textwidth}
        \centering
        \includegraphics[width=\linewidth]{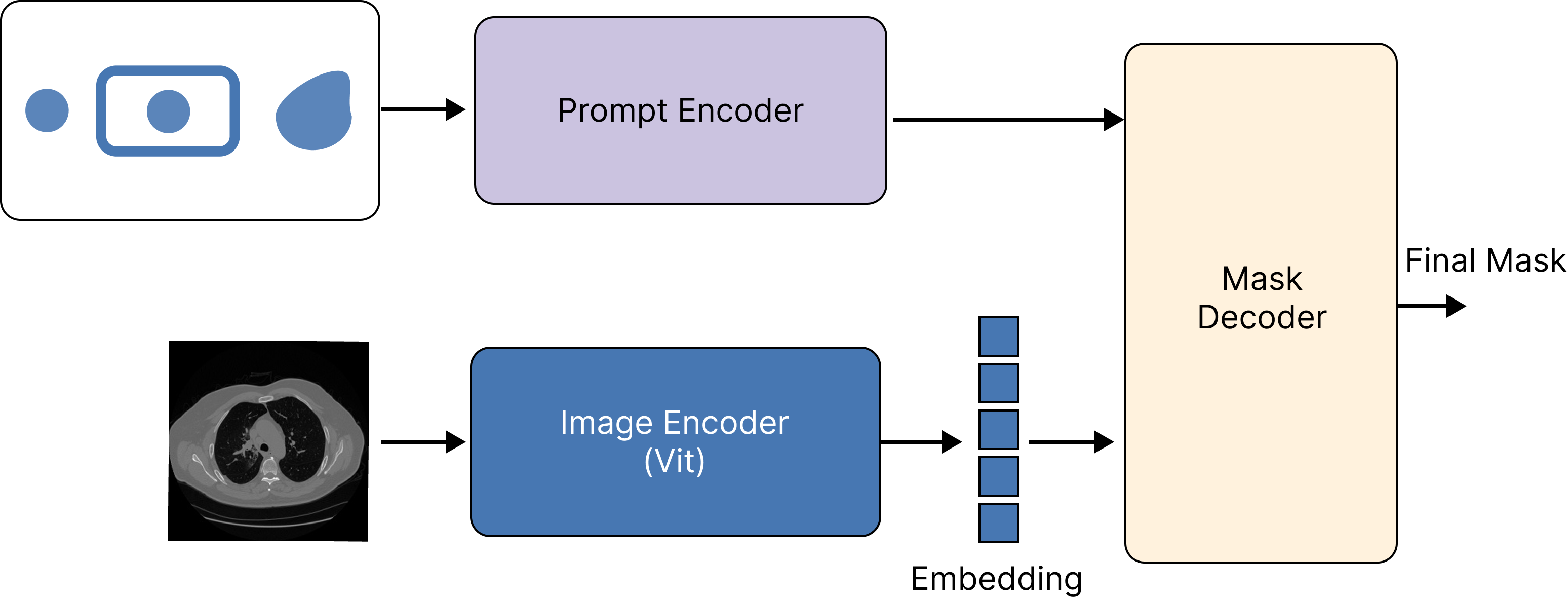}
        \caption{SAM}
        \label{fig:sam}
    \end{subfigure}
    \hfill
    \begin{subfigure}[b]{0.48\textwidth}
        \centering
        \includegraphics[width=\linewidth]{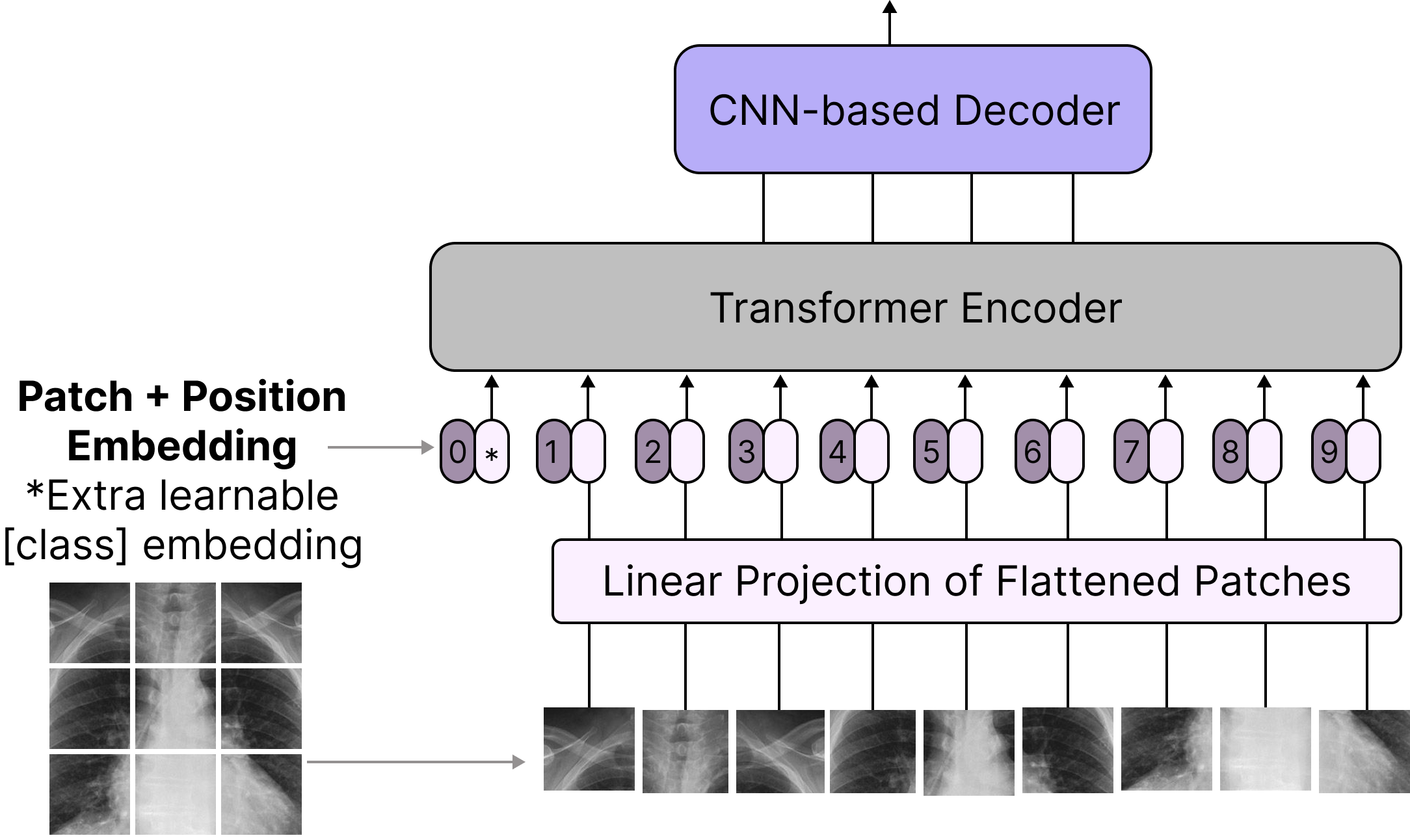}
        \caption{Hybrid (Transformers and CNN)}
        \label{fig:hybrid}
    \end{subfigure}
    \caption{Figure illustrates key architectural innovations in computer vision that have shaped FMs. (a) Vision Transformer (ViT) introduced scalable transformer-based image representations; (b) Masked Autoencoder (MAE) improved feature learning through masked image reconstruction; (c) SAM enabled promptable segmentation across diverse domains with minimal tuning; (d) Hybrid architectures combine CNNs and transformers to capture both local features and global context. Together, these advances have enhanced the scalability, generalization, and transferability of modern FMs across many visual tasks.}

    \label{fig:model-architecture}
\end{figure}

\subsubsection{Vision Transformer} \label{vit}
The ViT (Figure \ref{fig:vit}) represents a pivotal advancement in computer vision, adapting the transformer architecture originally developed for NLP to image analysis. Unlike CNNs, which extract features by applying learnable convolutional filters, ViT divides an image into fixed-size patches and treats each patch as an individual token. These tokens are then processed through self-attention layers, enabling the model to capture intricate relationships across the entire image \cite{ref19_vit}. This design facilitates global context modeling from the outset, but unlike CNNs, ViTs lack \textit{inductive biases} such as locality (nearby pixels are related) and translation equivariance (features are recognizable regardless of their location). Instead, ViTs must learn such priors directly from data, which becomes feasible when trained on large and diverse datasets. By foregoing these constraints, ViTs enable flexible and expressive representations that can surpass convolutional architectures given sufficient data \cite{ref7_swinunetr,ref19_vit}.

ViT has achieved SOTA performance on several computer vision benchmarks, and its impact is increasingly evident in medical imaging. Recent studies have demonstrated its effectiveness in tasks such as tumor classification and multi-organ segmentation \cite{ref6_unetr,ref50_transmed}. However, the model’s strong reliance on massive labeled datasets presents a significant limitation, especially in medical imaging domains where annotated data is often scarce and expensive. In contrast, CNNs benefit from image-specific inductive biases and typically require less data to achieve good performance, making ViTs comparatively data-hungry \cite{ref19_vit}. To address this limitation, researchers have proposed hybrid architectures that combine the spatial inductive advantages of CNNs with the global modeling capabilities of transformers \cite{ref6_unetr, ref7_swinunetr}. Additionally, self-supervised and transfer learning strategies have been explored, enabling ViTs to be pretrained on large-scale unlabeled datasets and then fine-tuned on limited labeled data. These approaches improve sample efficiency and enhance performance in data-constrained medical imaging scenarios.

\subsubsection{Masked Autoencoders}\label{mae}
 Masked Autoencoders (MAE) \cite{ref51_mae} extend the transformer-based architecture of ViT by introducing an efficient and scalable approach for learning transferable visual representations without labeled data. The core architectural innovation of MAE is its asymmetric encoder-decoder structure as shown in Figure \ref{fig:mae}. Unlike conventional autoencoders, where encoder and decoder capacities are often similar, MAE’s encoder processes only the small subset of visible image patches (typically 25\%), while the majority (75\%) are masked. The lightweight decoder is then tasked to reconstruct the missing pixels from latent representations and mask tokens in a self-supervised manner. This approach not only accelerates training (up to threefold faster), but also improves accuracy and reduces computational demands \cite{ref51_mae}. Importantly, MAE challenges the traditional reliance on fully supervised pretraining, showing that simple pixel reconstruction from masked inputs can yield rich and transferable visual representations. While these emergent representations have shown competitive performance across a wide range of downstream tasks, their effectiveness is closely tied to the choice of masking ratio and pretext task (see Section \ref{ssl}), and benefits are most pronounced when pretraining is conducted on large and diverse datasets \cite{ref114_simmim, ref26_ssl_mae_mia}. Furthermore, while MAE demonstrates strong generalization for natural images, adapting MAE to medical applications requires careful consideration of optimal masking ratios, patch sizes, reconstruction targets, decoder design, and masking strategies to ensure robust and transferable representations \cite{ref52_mim3d}. 

\subsubsection{Promptable and Interactive Models}
The development of promptable and interactive vision models represents a paradigm shift in how segmentation and annotation tasks are handled in both general and medical imaging domains. Traditional approaches such as U-Net \cite{ref5_unet},  typically require extensive retraining or fine-tuning to adapt to new object classes or anatomical structures, limiting their flexibility. In contrast, promptable models are explicitly designed to generalize across a wide range of tasks through user-provided prompts or queries, enabling on-the-fly adaptation without the need for retraining. 

A notable advancement is SAM, which uses MAE pretrained ViT as an image encoder, while prompts are processed through a dedicated prompt encoder. The combined embeddings are then fed to a mask decoder, which generates segmentation masks as shown in Figure \ref{fig:sam}. Trained on over one billion masks, SAM can interactively segment objects with minimal supervision, offering flexibility and efficiency far beyond conventional approaches.  It has also demonstrated strong zero-shot performance, often matching or surpassing fully supervised methods \cite{ref21_sam}. Crucially, the SAM framework redefines segmentation model design by decoupling model training from task-specific adaptation. This allows users to interactively guide the segmentation process, correct errors, or segment novel structures with minimal technical effort. As a result, it accelerates annotation and reduces reliance on expert labeling, which is especially beneficial in the medical domain. However, the direct adaptation of SAM to medical imaging results in suboptimal performance, primarily due to the model's training on natural images with clear RGB textures and object boundaries. As a result, SAM lacks the domain-specific priors needed for accurate medical segmentation \cite{ref_40_sammed2d, ref31_medsam}. Recent work demonstrates that targeted adaptation strategies are essential to bridge this performance gap and make promptable models clinically relevant \cite{ref53_medsa, ref54_SAMed, ref55_autosam}

\subsubsection{Multimodal Models}
FMs that jointly learn from multiple modalities, represent a key advancement in computer vision. By aligning visual representations with natural language, these models enable rich cross-modal reasoning, semantic understanding, and robust zero-shot transfer, setting new performance benchmarks across a range of tasks in both general and medical domains \cite{ref22_clip, ref32_medclip, ref28_bioclip}. A prominent example is CLIP, which employs a dual-encoder structure, where an image encoder (often a ViT or ResNet) and a text encoder (typically a transformer-based language model) are trained simultaneously on vast collections of image–text pairs. By maximizing the cosine similarity between matched image–text pairs and minimizing it for mismatched pairs, it ensures that related visual and textual representations are mapped closely together \cite{ref22_clip}.

CLIP challenges the conventional paradigm of task-specific supervised learning, advancing a more general and scalable FM framework built on contrastive multimodal pretraining. However, its reliance on web-scale, noisy image–text data introduces inherent biases, raising concerns about robustness, fairness, and domain generalization, particularly in high-stakes fields like medical imaging \cite{ref22_clip}.  A key challenge in this context is the prevalence of false negatives in contrastive learning. Although images and reports from different patients are typically treated as negative pairs, they may in fact describe the same symptoms or findings. This practice introduces noisy supervision and can confuse the model, undermining the quality of learned representations \cite{ref32_medclip}. These limitations necessitate domain-aware adaptations like MedCLIP, which mitigate false negatives and leverage unpaired data through knowledge-driven supervision.  Furthermore, while CLIP enables flexible downstream adaptation, its dual-encoder architecture imposes limitations on fine-grained spatial reasoning and dense prediction tasks \cite{ref21_sam, ref31_medsam}. These shortcomings have motivated architectural innovations, including SAM and hybrid multimodal frameworks, which integrate pixel-level supervision or unified encoder–decoder designs to support more detailed and clinically relevant predictions.

\subsubsection{Hybrid and Hierarchical Models}
As vision FMs continue to evolve, it has become increasingly clear that neither convolutional nor transformer-based architectures alone are universally optimal for the diverse challenges of imaging tasks. This recognition has fueled the development of hybrid and hierarchical models that combine the strengths of both paradigms to address their respective limitations.

\textbf{Hybrid models} integrate CNNs with transformer layers as shown in Figure \ref{fig:hybrid}, leveraging inductive biases of CNNs for local feature extraction and translation invariance, while exploiting transformers’ capacity to capture global context through self-attention mechanisms \cite{ref6_unetr}. Popular examples such as TransUNet \cite{ref56_transunet}, Swin-Unet \cite{ref57_swin_unet}, and UNETR \cite{ref6_unetr} have demonstrated that these combinations can yield superior performance in tasks like organ segmentation and volumetric image analysis, particularly in data-limited regimes where pure transformer models tend to underperform \cite{ref56_transunet,ref57_swin_unet}. Hybrid architectures also offer a practical means of scaling to high-resolution or 3D medical data, reducing the computational burden and data requirements associated with end-to-end transformer models.

\textbf{Hierarchical transformers} advance this concept further by organizing attention mechanisms across multiple spatial scales or resolutions as exemplified by Swin Transformer \cite{ref20_swin}. Swin introduces a window-based self-attention mechanism that operates within local windows and shifts these windows across the image, enabling efficient and scalable modeling of both local and global dependencies \cite{ref20_swin, ref7_swinunetr}. This hierarchical structure provides a strong inductive bias for integrating information at multiple levels, demonstrating strong performance with top-1 accuracy of 87.3\% on the ImageNet-1K benchmark \cite{ref20_swin}. In the medical domain, Swin-based variants such as Swin-UNETR have been increasingly adopted for volumetric image analysis and have established new standards for accuracy and efficiency in medical image segmentation \cite{ref7_swinunetr, ref57_swin_unet}.

Despite their promise, hybrid and hierarchical models introduce new complexities in both model design and training. The need to carefully balance convolutional and attention-based components often complicates architectural decisions and hyperparameter optimization \cite{ref6_unetr, ref56_transunet}. As a result, model development and optimization become more resource-intensive, often requiring extensive experimentation to achieve optimal performance.

\subsection{Pretraining Paradigms}
A defining feature of FMs is not only their architecture but also the paradigm used for large-scale pretraining, which determines the generalizability and transferability of learned representations. Although many pretraining strategies have emerged, we focus on two dominant approaches, supervised and self-supervised pretraining, which have been most influential in shaping how FMs acquire transferable representations for medical imaging.
\subsubsection{Supervised Pretraining}
Supervised pretraining has long been the preferred pretraining strategy, in which models are trained on large-scale datasets with human-provided labels, enabling models to learn robust, class-discriminative features. In this approach, models are trained on pairs of images $x$ and discrete class labels $y$, learning to predict the correct label for each image by minimizing the cross-entropy loss:

\begin{equation}
L = -\frac{1}{N} \sum_{i=1}^N \sum_{c=1}^C \mathbf{1}\{y_i = c\} \log p_\theta(y_i = c \mid x_i),
\end{equation}
where $N$ is the number of training samples, $C$ is the number of classes, $\mathbf{1}\{y_i = c\}$ is an indicator for the true class, and $p_\theta$ is the predicted probability. This approach has enabled models such as ViT, when pretrained on large datasets like ImageNet-21k or JFT-300M, to achieve strong performance across multiple image recognition benchmarks \cite{ref19_vit}. While supervised pretraining excels at learning transferable features, it is fundamentally tied to the availability of large, diverse, and accurately labeled datasets. In domains like medical imaging, where expert annotations are limited, this poses a significant challenge \cite{ref58_ssl_image_context_restoration}. It is also sensitive to the quality of annotations, as any biases present in training labels such as imbalanced label or systematic errors can be amplified in the learned representations. This can lead to poor generalization, particularly when models are deployed in domains that differ semantically or structurally from their training data. This often necessitates extensive domain-specific fine-tuning \cite{ref59_big_ssl}. Moreover, the rigid nature of supervised pretraining can make models susceptible to spurious correlations, overfit to dataset-specific artifacts, and vulnerable to adversarial attacks \cite{ref60_ssl_contrastive_generative, ref61_survey_contrastive_ssl}. 

To address these limitations, recent advances propose several augmentation strategies. Techniques like active learning, which prioritize labeling high-uncertainty samples \cite{ref142_active_learning}, or self-supervised or semi-supervised pretraining can reduce the annotation burden. Additionally, synthetic data generation using generative adversarial networks (GANs) or diffusion models offers a promising avenue to augment scarce labeled datasets, particularly in medical imaging \cite{ref33_comprehensive_survey_fm,ref10_fm,ref34_generalist_fm}. Moreover, hybrid pretraining paradigms that combine supervised and SSL help models capture both class-specific and general features, thereby mitigating overfitting to dataset-specific artifacts. Despite these challenges, supervised learning remains essential as the final fine-tuning stage for adapting models pretrained with self-supervised or multimodal approaches, providing the necessary task and domain-specific information required for reliable clinical performance. In the context of FMs, supervised pretraining represents both a historical foundation that enabled early advances and a complementary paradigm that still plays a critical role in guiding downstream adaptation.

\subsubsection {Self-Supervised Pretraining}\label{ssl}
In contrast, SSL has emerged as a critical paradigm underpinning the scalability and adaptability of modern FMs by directly addressing the persistent challenges of expert-annotated data. By leveraging large pools of unlabeled data, SSL enables models to learn robust, transferable representations without reliance on extensive manual labeling
\cite{ref60_ssl_contrastive_generative, ref62_simclr}. This approach is particularly suited for the medical domain, where privacy constraints, specialized expertise requirements, and the time-consuming nature of annotation often limit the scale and diversity of labeled datasets. SSL is typically structured around two distinct but interconnected tasks: the \textit{pretext task} and the \textit{downstream task}. In the pretext phase, the model learns to solve an automatically constructed task (e.g., inpainting, rotation prediction, or masked modeling), which forces it to capture generalizable features from the data  \cite{ref1_3d_ssl_methods_medical}. For a set of unlabeled images
$\mathcal{D}_{\text{unlabeled}} = \{ x_i \}_{i=1}^N,$
pseudo-labels ${y_i}^{\text{pretext}}$ are generated directly from each image, and the self-supervised model $f_{\mathrm{SSL}}$ (parameterized by $\theta$) is optimized using an appropriate loss function $\ell$ (e.g., cross-entropy, MSE, contrastive loss):
\begin{equation}
\theta^* = \arg\min_{\theta} \; \mathcal{L}_{\text{pretext}}(\theta).
\end{equation}
Once pretext task training is complete, the learned parameters $(\theta^*)$ serve as an effective initialization for downstream clinical tasks. The downstream model is then fine-tuned on a small set of annotated examples, or used as a fixed feature extractor, optimizing a supervised loss $\ell_{\mathrm{sup}}$ over the labeled dataset
$\mathcal{D}_{\text{labeled}} = \left\{ (x_j, y_j) \right\}_{j=1}^M:$
\begin{equation}
\phi^* = \arg\min_{\phi} \; \frac{1}{M} \sum_{j=1}^M \ell_{\mathrm{sup}}\big( f_{\mathrm{down}}(x_j; \theta^*, \phi), \; y_j \big).
\end{equation}

While SSL is already transforming MIA (see Section \ref{ssl_pretraining}), challenges remain in understanding the optimal design of pretext tasks and ensuring robust transfer across modalities and institutions. Although generic pretext tasks often yield broad transferable features, recent research suggests that customizing SSL tasks to exploit anatomical structure, clinical semantics, or imaging physics can further enhance downstream performance \cite{ref1_3d_ssl_methods_medical, ref25_ssl_swin_trans}. Another significant limitation is the interpretability of learned representations, which can hinder clinical acceptance. To address these issues, ongoing research has focused on developing domain-adapted pretext tasks that better align with clinical objectives, hybrid methods that incorporate both labeled and unlabeled data, and standardized benchmarking initiatives to ensure robust and fair evaluation. Notably, SSL-pretrained models have been shown to rival, and in data-scarce scenarios, even outperform fully supervised counterparts on tasks such as disease classification, tumor or organ segmentation, and anomaly detection \cite{ref26_ssl_mae_mia,ref25_ssl_swin_trans,ref59_big_ssl}. 
This establishes SSL as a cornerstone for FMs in medical imaging, as it directly supports the core objective of exploiting large unlabeled datasets to learn generalizable representations that can be efficiently adapted with minimal supervision.

%% file: 04_adaptations.tex
\section{Adaptation Techniques for Medical Image Analysis} \label{sec4:adaptations}
In this section, we discuss key adaptation strategies for adapting FMs for MIA. These include fine-tuning approaches, parameter-efficient adaptations, self-supervised pretraining, and multimodal or cross-modal techniques, each addressing different challenges of domain shift, data scarcity, and clinical applicability.
\subsection{Fine-tuning Approaches}
Fine-tuning adapts a pretrained model to a specific downstream task by updating some or all of its parameters using a smaller, task-specific dataset. This strategy allows models to leverage broad visual knowledge while specializing in the unique distribution and diagnostic features of medical data \cite{ref25_ssl_swin_trans, ref63_cnn_full_or_fine}. While each fine-tuning strategy offers specific advantages such as computational efficiency, stability, or improved task adaptation, these benefits often come with trade-offs in terms of performance, resource usage, or generalizability. To illustrate these differences, Table~\ref{tab:finetuning_comparison} presents a comparative overview of fine-tuning strategies evaluated on the LUNA25 lung nodule malignancy dataset. It highlights variations in trainable parameters, memory usage, training time, and validation performance.

\subsubsection{Linear probing (Head-only Training)}
Linear probing represents the simplest and most computationally efficient form of transfer learning for deep neural networks. In this approach, the pretrained backbone is kept entirely frozen, and only a newly added linear head, typically a single fully connected layer is trained on the downstream task \cite{ref64_ft_distor_feat}. Also known as head-only training, this method eliminates any gradient updates to the feature extractor, relying solely on the quality of the learned representations.  While not a fine-tuning method per se, linear probing is widely used as a baseline diagnostic tool to assess the quality of the learned representations before any task-specific adaptation.

The primary motivation behind linear probing is its efficiency. It requires minimal computational resources, eliminates the risk of \textit{catastrophic forgetting} (the loss of previously learned knowledge), minimizes overfitting, and enables rapid adaptation to new tasks \cite{ref1_3d_ssl_methods_medical, ref10_fm, ref64_ft_distor_feat,ref148_experience_replay}. This approach has become a standard diagnostic tool in both research and clinical AI, providing a low-cost, rapid benchmark to assess the effectiveness of pretraining.  For instance, linear probing has been used to evaluate the quality of self-supervised representations for 3D medical imaging across multiple modalities \cite{ref1_3d_ssl_methods_medical}, and to benchmark the transferability of large-scale self-supervised models on a range of medical image classification tasks \cite{ref59_big_ssl}. Within the FM context, linear probing is particularly suitable for rapidly benchmarking how well generic representations transfer to the specialized medical domain.

Despite its computational appeal, linear probing is inherently limited in expressivity and adaptability. Since only the final linear layer is updated, the model cannot re-align or adapt internal representations to new or significantly shifted domains. This limitation becomes evident in complex or highly specialized applications, such as fine-grained tumor segmentation or rare disease detection, where partial or full fine-tuning of the backbone is often required for optimal performance \cite{ref64_ft_distor_feat,ref65_transfusion}. In our experiments (Table \ref{tab:finetuning_comparison}), while linear probing incurs the lowest computational cost, it also yields the lowest validation AUC (59.5\%) among all strategies evaluated. Furthermore, success with linear probing on well-curated benchmarks may overstate the robustness of frozen representations, potentially masking critical weaknesses when deployed in more heterogeneous or clinically realistic environments.

\subsubsection{Full Fine-tuning}
Full fine-tuning involves updating all parameters of a pretrained model when adapting it to a specific downstream task \cite{ref64_ft_distor_feat}. This strategy is motivated by the need to maximize adaptability, especially in scenarios where the target domain (e.g., CT or MRI) diverges substantially from the data distribution of the original pretraining datasets, such as ImageNet \cite{ref19_vit, ref65_transfusion}. Updating all weights maximizes the model’s adaptability to capture domain-specific features that may not align with the original pretraining data distribution.
When sufficient labeled data and computational resources are available, full fine-tuning often yields the highest performance. For example, Liang and Zheng \cite{ref66_tl_residal_net} applied full fine-tuning by initializing the model with weights pretrained on the ChestXray14 dataset and then updating all parameters on the pediatric pneumonia dataset, resulting in improved accuracy of 90.5\% and F1-score of 92.7\% outperforming other popular CNN architectures. Similarly, end-to-end adaptation of ViT on the four chest X-ray datasets improved classification metrics by 3–5\% (recall and F1-score respectively) over all ViT variants \cite{ref67_ivevit}.  

However, the real-world applicability of full fine-tuning is particularly constrained when adapting FMs to medical imaging, where data scarcity and quality issues can make full model adaptation prone to severe overfitting \cite{ref68_peft_review}. Consistent with this, our experiments (Table \ref{tab:finetuning_comparison}) indicate that while full fine-tuning can achieve strong performance, it incurs substantial computational overhead, requiring approximately 20\% more memory than alternative approaches. Moreover, updating all weights also increases the risk of catastrophic forgetting, especially when models are sequentially adapted to multiple tasks \cite{ref10_fm, ref69_unilm_ft_text}, raising concerns about consistency and
reliability for clinical applications.  Some evidence suggests that full fine-tuning may even underperform compared to linear probing or partial adaptation when pretrained features are already highly informative, or when there is a substantial distribution shift between the pretraining and target domains \cite{ref64_ft_distor_feat}. These challenges have motivated researchers to optimize full fine-tuning through techniques like regularization, which employ methods that help the model to prevent overfitting or incorporate partial and parameter-efficient fine-tuning strategies to minimize computational overhead.
\begin{table}[!htbp]
\centering
\begin{threeparttable}
\caption{Comparative evaluation of four fine-tuning strategies on the LUNA25 lung nodule malignancy dataset. Models were trained for 20 epochs on 3D nodule patches (\(N=6{,}163\)) with associated clinical features, using an NVIDIA A100 (80GB) GPU. The results illustrate trade-offs between computational efficiency and predictive performance.}
\label{tab:finetuning_comparison}
\renewcommand{\arraystretch}{1.15}
\setlength{\tabcolsep}{3pt}
\begingroup
\fontsize{9}{11}\selectfont
\begin{tabularx}{\textwidth}{%
    l                                  
    >{\raggedleft\arraybackslash}X     
    >{\raggedleft\arraybackslash}X     
    >{\raggedleft\arraybackslash}X     
    >{\raggedleft\arraybackslash}X     
    >{\raggedleft\arraybackslash}X     
    >{\raggedleft\arraybackslash}X     
}
\toprule
Strategy & Trainable Params (Avg) & Trainable Params (Final) & GPU Mem Allocated (GB) & Train Time (min) & Best Val AUC & Final Train Loss \\
\midrule
Linear probing     & \(131{,}329\)   & \(131{,}329\)   & \(0.193\) & \(25.76\) & \(0.595\) & \(0.305\) \\
Full fine-tuning   & \(46{,}286{,}401\) & \(46{,}286{,}401\) & \(0.723\) & \(73.09\) & \(0.900\) & \(0.130\) \\
Gradual unfreezing & \(36{,}367{,}067\) & \(46{,}264{,}321\) & \(0.604\) & \(60.65\) & \(0.875\) & \(0.142\) \\
Discriminative FT  & \(36{,}367{,}067\) & \(46{,}264{,}321\) & \(0.604\) & \(62.67\) & \(0.826\) & \(0.048\) \\
\bottomrule
\end{tabularx}
\endgroup
\end{threeparttable}
\end{table}

\subsubsection{Partial Fine-tuning}
Partial fine-tuning offers a pragmatic alternative to balance the trade-offs between the rigidity of linear probing and the risks of catastrophic forgetting in full fine-tuning. In this approach, only a selected subset of model parameters, often the later, more task-specific layers are updated for the new task, while the rest remain frozen \cite{ref68_peft_review}. This approach is motivated by the observation that earlier network layers tend to learn robust, general visual features, whereas later layers encode more specialized, domain- or task-specific information \cite{ref65_transfusion, ref76_how_transferable_dl}.
For all model parameters $\theta$ and $\theta_T \subset \theta$ is the subset targeted for adaptation, partial fine-tuning optimizes only $\theta_T$ (with $\theta_F$ frozen) by minimizing a task-specific loss function $L$:
\begin{equation}
\theta_T^* = \arg\min_{\theta_T} \frac{1}{N} \sum_{i=1}^{N} L\left(f(x_i; \theta_F, \theta_T),\; y_i\right).
\end{equation}

\textbf{Layer-wise fine-tuning (Gradual Unfreezing):} This is one of the most widely used partial fine-tuning approaches in which the earlier layers of a pre-trained network are initially kept frozen, while only the higher layers are progressively unfrozen and updated during adaptation \cite{ref69_unilm_ft_text}. This adaptation strategy preserves broad, generalizable features learned by early layers during large-scale pretraining, while allowing higher layers to efficiently adapt to the unique characteristics of downstream tasks. A popular example of this strategy was proposed by Felbo et al.\cite{ref70_chain_thaw} called \textit{chain-thaw}, which sequentially unfreezes and fine-tunes one layer at a time, allowing the model to adapt each layer with minimal overfitting. Howard and Ruder \cite{ref69_unilm_ft_text} extended this idea by introducing a strategy called \textit{gradual unfreezing}. In this approach, fine-tuning begins by unfreezing only the final layer, which is trained for one epoch. After each subsequent epoch, the next preceding layer is unfrozen, and all currently unfrozen layers are jointly fine-tuned.  This process continues until the entire network is unfrozen or performance converges. Such approaches help improve stability and transferability while reducing the risk of overfitting during fine-tuning.

Several studies have demonstrated the effectiveness of layer-wise fine-tuning across diverse medical imaging tasks and modalities. For instance, Tajbakhsh et al.\cite{ref63_cnn_full_or_fine} systematically evaluated layer-wise fine-tuning in four distinct medical applications, including radiology, cardiology, and gastroenterology. This method consistently matched or outperformed CNNs trained from scratch and proved more robust to varying training set sizes. Swati et al.\cite{ref71_block_wise_ft} adopted a block-wise fine-tuning strategy on T1-weighted CE-MRI benchmark dataset, achieving SOTA accuracy without relying on handcrafted features or extensive preprocessing. Hermessi et al.\cite{ref72_deep_tissue_sarcoma} demonstrated that layer-wise fine-tuning of pre-trained CNNs for soft tissue sarcoma classification in MRI substantially improved classification accuracy from 93\% to 98.7\%. Moreover, several other studies have reported comparable or even improved performance, such as BACH breast tissue dataset classification \cite{ref73_comp_ft_st}, breast cancer histopathological image classification \cite{ref74_layer_wise_breast_cancer}, and diabetic retinopathy fundus image classification \cite{re75_retinopathy}. 

Despite these benefits, several important considerations must be addressed when adopting this strategy. There is no universally optimal policy for determining how many layers to unfreeze or the precise schedule for doing so. These choices are highly dependent on the specific task and dataset, and typically require substantial empirical tuning. Moreover, while freezing early layers helps retain general features, it may also limit the model’s ability to adapt to domains with substantial distribution shifts, potentially resulting in suboptimal performance \cite{ref65_transfusion,ref69_unilm_ft_text}. For instance, if the pretraining and target domains are not well aligned, limiting adaptation to later layers may leave domain-specific artifacts or biases uncorrected in early representations \cite{ref76_how_transferable_dl}. This is particularly concerning in medical imaging, where subtle features in early layers can be clinically relevant. Furthermore, model performance can be sensitive to particular layers selected for adaptation and the sequence in which they are unfrozen. Future research must focus on developing adaptive unfreezing strategies and systematically benchmarking their effectiveness across diverse medical imaging tasks and modalities.

\textbf{Discriminative fine-tuning:} Originally proposed in the NLP by Howard and Ruder, this adaptation strategy assigns different learning rates to different layers of the model \cite{ref69_unilm_ft_text}. Unlike conventional approaches that uses a uniform learning rate across all layers, discriminative fine-tuning typically applies smaller learning rates to the early, general-purpose layers and progressively larger rates to deeper, more task-specific layers. This method enables the model to adaptively refine relevant features for the target domain while preserving stable, generalizable representations learned during pretraining \cite{ref69_unilm_ft_text, ref76_how_transferable_dl}. In the context of adapting FMs, this strategy offers a principled way to balance stability and flexibility, enabling effective domain adaptation with limited data. Mathematically, the model parameters $\boldsymbol{\theta}$ is divided into subsets $\{\boldsymbol{\theta}_1, \ldots, \boldsymbol{\theta}_L\}$, where $\boldsymbol{\theta}_l$ denotes the parameters of the $l$-th layer for a total of $L$ layers. During optimisation, each layer $l$ is updated via stochastic gradient descent (SGD) using its own learning rate $\eta^l$:
\begin{equation}
\theta_t^l = \theta_{t-1}^l - \eta^l \cdot \nabla_{\theta^l} J(\theta),
\end{equation}
where $\nabla_{\boldsymbol{\theta}_l} J(\boldsymbol{\theta})$ is the gradient of the loss with respect to that layer’s parameters \cite{ref69_unilm_ft_text}. This flexibility allows the deeper layers to adapt rapidly, while shallow layers remain stable, minimizing both catastrophic forgetting and overfitting,

A notable variant is \textit{Layer-wise Learning Rate Decay (LLRD)}, which starts with a relatively high learning rate for the topmost layer and decays it multiplicatively for each successive lower layer \cite{ref77_llrd}. Other variants such as custom grouping of layers into blocks (e.g., early, middle, late) with distinct rates \cite{ref71_block_wise_ft}, and hybrid strategies, which combine discriminative learning rates with gradual unfreezing of layers provide additional flexibility. In medical imaging, Adedigba et al. \cite{ref78_optimal_hyper} adopted discriminative fine-tuning strategy for COVID-19 chest X-ray classification, achieving fast convergence (peak accuracy in 20 epochs) and strong generalization on imbalanced datasets, with a validation accuracy of 96.8\%. Moreover, this approach has been increasingly adopted in other applications, such as breast cancer histopathological image classification \cite{ref74_layer_wise_breast_cancer} and tuberculosis detection~\cite{ref79_dl_tbclassification}. Despite its effectiveness, discriminative fine-tuning lacks a principled approach for assigning learning rates, often relying on empirical or heuristic methods that can be computationally expensive \cite{ref69_unilm_ft_text}. Moreover, the sensitivity of discriminative fine-tuning to hyperparameters and model depth can lead to suboptimal adaptation or instability. Addressing these challenges will require automating learning rate selection, developing theoretical foundations for layer-wise adaptation, and rigorously benchmarking discriminative fine-tuning against parameter-efficient alternatives.

\subsection{Parameter-Efficient Adaptations}
PEFT offers compelling solutions for robust and lightweight task adaptation by updating only a small subset of model parameters \cite{ref80_delta_tuning}. Unlike conventional fine-tuning that requires updating billions of parameters in large models like GPT-3, PEFT methods can achieve comparable performance by adjusting as little as 0.5\% of the parameters, yielding substantial efficiency gains \cite{ref80_delta_tuning, ref81_power_scale_peft}. This approach significantly reduces computational demands, lowers memory requirements, mitigates the risk of overfitting, and is well-suited for scenarios with limited labeled data. In this review, we categorize PEFT methods into four principal categories, addition-based, selection-based, reparameterization-based, and hybrid strategies as illustrated in Figure \ref{fig:peft-taxonomy} and comparatively summarized in Table~\ref{tab:peft_summary}.

\input{peft_taxonomy}

\subsubsection{Addition-Based Methods}\label{adition-based}
Addition-based PEFT strategies adapt large models by injecting new lightweight modules, such as adapters, prompts, or  side networks into the backbone architecture, enabling efficient and modular task adaptation with minimal computational overhead. These approaches provide substantial memory and speed advantages, since only a small fraction of additional parameters are trained while the majority of the model remains frozen \cite{ref80_delta_tuning, re87_scaling, ref68_peft_review}.

\textbf{Adapter-based tuning} \label{adapter-based} inserts small, trainable bottleneck modules called \textit{adapters} into each transformer block while keeping the backbone frozen. For input $h \in \mathbb{R}^d$, the adapter projects $h$ to a lower dimension $r \ll d$ via $W_d$, applies a non-linearity $f(\cdot)$, and projects back via up-projection $W_u$.
\begin{equation}
\label{eq:adapter}
\mathrm{Adapter}(h) = h+ W_u\, f(W_d h).
\end{equation}
As expressed in Eq.~\eqref{eq:adapter}, most of the pretrained representation \(h\) is preserved, 
while the trainable bottleneck layers \(W_d\) and \(W_u\) learn compact, task-specific adjustments. The residual connection stabilizes training and prevents catastrophic forgetting, while the bottleneck (\(r \ll d\)) drastically reduces the number of additional parameters. As a result, adapters achieve efficient fine-tuning with minimal trainable parameters, typically only 0.5–8\% of the full model \cite{ref53_medsa,ref80_delta_tuning}.

\begin{table}[!htbp]
\centering
\begin{threeparttable}
\caption{Overview of adapter-based PEFT techniques, including their core mechanisms, parameter efficiency, computational trade-offs, and key features.}
\label{tab:adapter_peft_summary}
\renewcommand{\arraystretch}{1.15}
\setlength{\tabcolsep}{3pt}
\begingroup
\fontsize{9}{11}\selectfont
\begin{tabularx}{\textwidth}{%
    >{\raggedright\arraybackslash}p{2.0cm}  
    >{\raggedright\arraybackslash}p{3.0cm}  
    >{\raggedright\arraybackslash}p{2.2cm}  
    >{\raggedright\arraybackslash}p{2.7cm}  
    >{\raggedright\arraybackslash}X         
    >{\raggedright\arraybackslash}p{1.0cm}  
}
\toprule
Method & Core Mechanism & Params (\%) & Memory/Speed & Key Features & Ref \\
\midrule
Standard Adapter & Bottleneck layer in transformer blocks & \(0.5\text{--}8\%\) of full model & \(1.3\times\) (adapter) vs.\ \(9\times\) (full FT) & Within \(0.4\%\) of full FT accuracy, modular design, avoids forgetting & \cite{ref82_standard_adapter} \\

Compacter & Kronecker-summed low-rank hypercomplex adapters & \(\sim 0.047\%\) of full model & \(\downarrow 35\%\) memory, \(\downarrow 13\%\) time vs.\ FT & High performance with ultra-low param count, effective in low-resource tasks & \cite{ref83_compacter} \\

Parallel Adapter & ReLU-based adapters in parallel with FFN or attention & \(2.4\text{--}12.3\%\) of full model & More efficient than sequential & Outperforms sequential, scalable across multiple heads & \cite{ref84_unified_peft}\textsuperscript{†} \\

AdapterFusion & Attention-based fusion via query-key-value mixing & \(\sim 3.6\%\times N + 1\%\) & \(\uparrow 1.82\%\) over full FT & Two-stage tuning, strong in multi-task learning, avoids catastrophic forgetting & \cite{ref85_adapterfusion} \\
\bottomrule
\end{tabularx}
\endgroup
\begin{tablenotes}
\footnotesize
\item[†] The implementation of the parallel adapter method follows the architectural design proposed in \cite{ref84_unified_peft}.
\item[*] Reported results follow the respective publications, benchmarks, datasets, and backbone configurations may vary.
\end{tablenotes}

\end{threeparttable}
\end{table}

As shown in Table \ref{tab:adapter_peft_summary}, several adapter-based PEFT methods were first popularized in NLP, and these innovations have recently inspired adaptations in medical imaging. For instance, Med-SA \cite{ref53_medsa} adapts SAM to the medical domain by inserting lightweight adapters in both the encoder and decoder, enabling updates to only 2\% of parameters. This approach achieved superior performance when experimented across 17 medical segmentation tasks spanning multiple modalities. Similarly, Brain-Adapter \cite{ref86_brain_adapter} adapts CLIP for neurological disorder diagnosis by inserting lightweight adapters between frozen 3D brain MRI encoders and clinical text encoders, enabling robust multimodal alignment with minimal overhead. However, despite its lightweight adaptation, adapter-based methods are highly sensitive to adapter size and placement, with suboptimal configurations often leading to underfitting, particularly on complex tasks or small-scale datasets. Moreover, while adapters promote modularity and continual learning, they may also introduce architectural rigidity, requiring careful integration with the backbone \cite{ref82_standard_adapter, ref83_compacter, ref84_unified_peft}. In multimodal contexts, designing effective strategies for cross-modal adapter fusion remains an open challenge and an active area of research.

\textbf{Prompt Tuning}\label{prompt-tuning}  enables efficient adaptation of large models by prepending or injecting a small set of learnable tokens, also known as \textit{soft prompts} to the input. These tokens are trained jointly with the model on a downstream task, enabling adaptation by updating only these prompt vectors while keeping the model’s core parameters frozen \cite{ref81_power_scale_peft}.  This approach can reduce tunable parameters to less than 0.02\% of the total model \cite{re87_scaling}, making it highly attractive for very large vision transformers and FMs. In MIA, the PUNETR framework demonstrated that updating only 0.51\% of parameters was sufficient to narrow the gap with full fine-tuning to 7.81 percentage points on TCIA/BTCV and 5.37–6.57 points on subsets of TotalSegmentator in mean DSC \cite{ref88_prompt_tuning1}. Similarly, DVPT achieved strong gains with just 0.5\% trainable parameters, surpassing full fine-tuning by +2.20\% in Kappa score on a classification benchmark, while reducing labeled data requirements by up to 60\% and storage costs by 99\% of ViT-B/16 \cite{ref89_prompt_tuning}. However, its effectiveness diminishes in low-resource scenarios, where smaller models or limited data can hinder optimization. Empirical studies suggest that while soft prompts can achieve competitive performance, they often converge more slowly than full fine-tuning or adapter-based approaches, requiring more training iterations to reach similar accuracy \cite{ref80_delta_tuning, ref91_revisiting_peft}.

\textbf{Prefix tuning}\label{prefix-tuning} provides a variant of prompt-based adaptation by introducing a sequence of continuous, task-specific vectors called \textit{prefixes}, not only at the model’s input, but throughout the internal layers. These prefixes are typically prepended to the key and value matrices in self-attention modules, enabling the injection of task-relevant information without modifying the model’s core parameters. By isolating task-specific information within compact prefix vectors, the method supports scalable and modular deployment across tasks, enabling multi-task learning, personalization, and efficient switching between tasks without parameter sharing or interference. Empirical studies in NLP have shown that prefix tuning can match or even surpass the performance of full fine-tuning in high-resource settings \cite{ref92_prefix_tuning}. While prefix tuning has been applied to medical task like visual question answering \cite{ref132_prefix} and report summarization \cite{ref141_radadapt}, its effectiveness hinges on critical design choices, such as the optimal prefix length and the specific transformer layers targeted for prefix injection. These hyperparameters significantly affect learning dynamics and model performance, especially under data-scarce conditions. Prefix tuning may converge more slowly or require careful hyperparameter tuning compared to other PEFT techniques, particularly in low-data or time-constrained settings \cite{re87_scaling, ref80_delta_tuning, ref83_compacter}.

\textbf{Side tuning} \label{side-tuning} introduces a lightweight, task-specific side network to the model, combining the outputs of the backbone and side network during inference. Unlike methods that modify the backbone or insert intermediate modules, side tuning enables full preservation of the pretrained model, reducing risks of catastrophic forgetting, overfitting, and task interference \cite{ref93_side_tuning}. This makes it well-suited for incremental learning and multi-task deployment, where previously learned capabilities must remain intact. Despite its demonstrated efficacy in NLP and vision-language tasks, side tuning remains unexplored in MIA. The effectiveness of side tuning is tightly coupled to design choices in the side network architecture and the fusion strategy used to combine outputs. Underpowered side networks may fail to capture clinically relevant domain shifts, while overly complex designs can erode the efficiency gains and introduce training instability \cite{ref93_side_tuning, ref94_peft_for_vision}. Hence, realizing its full potential in medical imaging requires principled exploration of architecture–fusion trade-offs, adaptive capacity control, and data-efficient training strategies.

\subsubsection{Selection-Based Methods}\label{selection-based}
Selection-based PEFT presents a minimalist yet effective alternative to both full fine-tuning and addition-based strategies. Instead of introducing new trainable components, it optimizes a carefully chosen subset of the existing model parameters, leaving the majority of the network unchanged \cite{ref94_peft_for_vision, ref95_peft_survey}. This approach leverages the insight that not all parameters contribute equally to task adaptation, and targeting key parameters enables efficient specialization with minimal computational and memory requirements \cite{ref82_standard_adapter}.
For the model parameters denoted as $\boldsymbol{\theta} = \{\theta_1, \theta_2, \ldots, \theta_n\}$ and binary mask $\mathbf{M} = \{m_1, m_2, \ldots, m_n\}, \quad \text{where } m_i \in \{0, 1\}$ determines whether a parameter \( \theta_i \) is trainable (\( m_i = 1 \)) or frozen (\( m_i = 0 \)). During training, gradients are updated only for the selected parameters:
\begin{equation}
\theta_i \leftarrow \theta_i - \eta \cdot \frac{\partial \mathcal{L}}{\partial \theta_i} \quad \text{if } m_i = 1,
\end{equation}
where \( \eta \) is the learning rate and \( \frac{\partial \mathcal{L}}{\partial \theta_i} \) is the gradient of the loss with respect to \( \theta_i \). This mechanism ensures that adaptation is both sparse and targeted, making it highly scalable for large models.
In practice, selection-based methods fall into two broad categories: (1)
 \textbf{heuristic-based approaches} like BitFit \cite{ref97_bitfit}, which update only specific parameter types (e.g., biases), and (2) \textbf{data-driven methods} like Diff Pruning \cite{ref99_diff_pruning} and Fish Mask \cite{ref98_fistmask}, which learn sparse masks using information-theoretic or gradient-based metrics. A comparative summary of their core mechanisms, parameter efficiency, and empirical findings is provided in Table~\ref{tab:selection_peft_summary}.

In the medical domain, Dutt et al.\cite{ref96_peft_mia} evaluated BitFit on six medical classification datasets, and found that despite updating less than 0.1\% of parameters, it outperformed full fine-tuning on BreastUS and performed comparably on Pneumonia CXR, demonstrating its potential for resource-constrained clinical settings. Methods like BitFit are particularly impressive in small-to-medium datasets, making them well-suited for medical applications where labeled data is often scarce  \cite{ref97_bitfit}. However, the broader adoption of selection-based methods in medical imaging remains limited. A key barrier is the difficulty of identifying suitable masking strategies, as mask selection is highly sensitive to task complexity and data distribution. This sensitivity can lead to training instability or degraded performance, underscoring the need for more systematic, data-driven approaches to parameter selection \cite{re87_scaling}.
\begin{table}[!htbp]
\centering
\begin{threeparttable}
\caption{Overview of representative selection-based PEFT methods, including their core mechanisms, parameter efficiency, computational trade-offs, and key features.}
\label{tab:selection_peft_summary}
\renewcommand{\arraystretch}{1.2}
\setlength{\tabcolsep}{3pt}
\begingroup
\fontsize{9}{11}\selectfont
\begin{tabularx}{\textwidth}{%
    >{\raggedright\arraybackslash}p{1.5cm}  
    >{\raggedright\arraybackslash}X         
    >{\raggedright\arraybackslash}p{1.6cm}  
    >{\raggedright\arraybackslash}p{3.0cm}  
    >{\raggedright\arraybackslash}X         
    >{\raggedright\arraybackslash}p{1.4cm}  
}
\toprule
Method & Core Mechanism & Params (\%) & Memory/Speed & Key Features & Ref \\
\midrule
BitFit & Update only bias terms of the model & \(0.03\text{--}0.09\%\) & Minimal overhead & Competitive with full FT on small-to-medium datasets & \cite{ref97_bitfit} \\

FISH Mask & Pre-compute fixed sparse mask using top-\(k\) Fisher info, update only selected weights & \(0.08\text{--}0.5\%\) & Memory and communication efficient & Matches or exceeds full FT at \(0.5\%\), avoids new parameters, supports distributed training & \cite{ref98_fistmask} \\

Diff Pruning & Learn sparse task-specific ``diff'' vector with \(L_0\)-norm relaxation & \(0.5\%\) & Memory efficient; more training cost than BitFit & Matches full FT on GLUE at \(0.5\%\), supports continual and on-device learning & \cite{ref99_diff_pruning} \\
\bottomrule
\end{tabularx}
\begin{tablenotes}
\footnotesize
\item[*] Reported results follow the respective publications; benchmarks, datasets, and base model configurations may vary.
\end{tablenotes}
\endgroup
\end{threeparttable}
\end{table}

\subsubsection{Reparameterization-Based Methods}\label{reparameterization-based}
Reparameterization-based PEFT methods represent a class of techniques that redefine the fine-tuning process by transforming the parameterization of neural networks. Unlike selection-based or addition-based approaches, this approach introduces a small set of trainable parameters that modify the behavior of a frozen pretrained model without altering its core architecture or weights \cite{ref68_peft_review, ref95_peft_survey, ref94_peft_for_vision}. A core idea underpinning these methods is that task-specific information often lies in a lower-dimensional subspace rather than the full parameter space \cite{ref80_delta_tuning,ref100_lora}. This assumption led to techniques such as Low-Rank Adaptation (LoRA)\cite{ref100_lora}, which freezes the original weights \( W_0 \in \mathbb{R}^{d \times k} \) and introduces trainable low-rank matrices \( A \in \mathbb{R}^{d \times r} \) and \( B \in \mathbb{R}^{r \times k} \), such that the adapted weight is expressed as:
\begin{equation}
W = W_0 + \Delta W = W_0 + AB,
\end{equation}
where the rank \( r \ll \min(d, k) \). During training, only \( A \) and \( B \) are updated while \( W_0 \) remains fixed. This allows for efficient fine-tuning with a significantly reduced number of trainable parameters. However, in the standard LoRA approach, a fixed rank is applied uniformly across all layers. Determining the optimal rank typically requires extensive tuning, and this uniform rank allocation can lead to suboptimal task adaptation \cite{ref101_adalora,ref102_dylora}.
To address these limitations, several LoRA variants have been proposed, each offering enhancements in rank allocation, adaptability, or efficiency, as outlined in Table \ref{tab:reparameterized_peft_summary}.

In MIA, LoRA consistently outperformed full fine-tuning for ViT-based classification tasks, especially in low-data regimes, achieving up to a 6\% absolute F1-score gains while tuning fewer than 0.2\% of model parameters. These gains were most pronounced with larger ViT backbones and datasets with limited labeled samples, such as BreastUS \cite{ref96_peft_mia}. Moreover, LoRA fine-tuning of the SAM image encoder led to a 13.93\% absolute increase in average DSC for multi-organ segmentation compared to only tuning the mask decoder \cite{ref54_SAMed}. In neuroimaging, LoRA improved classification accuracy by 1.3\%–13.5\% across multiple Alzheimer’s and dementia datasets \cite{ref105_medblip}. Despite these gains, current methods still face design limitations, including the use of the same adaptation strategy across all network layers, without considering the unique roles and representational needs of each layer. In addition, once the structure for adaptation is chosen, it usually remains fixed during training, leaving little room for later adjustments \cite{ref101_adalora, ref102_dylora}. These challenges suggest a need for more flexible and layer-aware approaches that can better balance model expressivity and efficiency. While some recent research such as AdaLoRA \cite{ref101_adalora} and DyLoRA\cite{ref102_dylora} has started to address these issues, finding the right balance between adaptability, stability, and parameter efficiency remains an open research challenge.

\begin{table}[!htbp]
\centering
\begin{threeparttable}
\caption{Overview of reparameterization-based PEFT techniques including their core mechanisms, parameter efficiency, computational trade-offs, and key features.}
\label{tab:reparameterized_peft_summary}
\renewcommand{\arraystretch}{1.22}
\setlength{\tabcolsep}{3pt}
\begingroup
\fontsize{9}{11}\selectfont
\begin{tabularx}{\textwidth}{%
  >{\raggedright\arraybackslash}p{1.8cm}  
  >{\raggedright\arraybackslash}X         
  >{\raggedright\arraybackslash}p{1.7cm}  
  >{\raggedright\arraybackslash}p{2.8cm}  
  >{\raggedright\arraybackslash}X         
  >{\raggedright\arraybackslash}p{1.0cm}  
}
\toprule
Method & Core Mechanism & Params (\%) & Memory/Speed & Key Features & Ref \\
\midrule
LoRA & Inject low-rank matrices into weight updates & \(0.01\text{--}0.1\%\) & \(\sim 3\times\) lower vs.\ full FT, faster training & Efficient and orthogonal, enables task switching with minimal overhead & \cite{ref100_lora} \\

AdaLoRA & Dynamically allocate parameter budget among weight matrices & \(0.08\text{--}0.2\%\) & \(\uparrow\,11\text{--}16\%\) time, \(\uparrow\,0.6\%\) memory vs.\ LoRA & Dynamic SVD-based rank scheduling\textsuperscript{†} & \cite{ref101_adalora} \\

DyLoRA & Train LoRA modules across a dynamic range of ranks (instead of a single rank) & \(0.02\text{--}0.11\%\) & Up to \(7\times\) faster than LoRA & Search-free, inference-time rank selection, robust across larger range of ranks & \cite{ref102_dylora} \\

KronA & Use Kronecker product in parallel to weight matrices & \(0.07\%\) & \(\uparrow\,6.5\%\) train, \(\uparrow\,21\%\) inference vs.\ LoRA & Improved accuracy at the cost of increased inference latency & \cite{ref103_krona} \\
\bottomrule
\end{tabularx}
\begin{tablenotes}
\footnotesize
\item[†] \textit{SVD} (Singular Value Decomposition) extracts low-rank structure for adaptive capacity allocation.
\item[*] Reported results follow the respective papers; benchmarks, datasets, and base models may vary.
\end{tablenotes}
\endgroup
\end{threeparttable}
\end{table}

\subsubsection{Hybrid PEFT}\label{hybrid-methods}
While individual PEFT methods have shown strong performance, their inherent trade-offs limit their ability to fully address the complex requirements of real-world applications. As a result, hybrid PEFT offers an alternative by combining complementary strengths to deliver more robust and versatile adaptation, particularly in challenging domains like medical imaging. These approaches explicitly integrate two or more PEFT methods within a single model to improve generalization and task adaptability \cite{ref68_peft_review}. For instance, MAM adapter strategically applies prefix tuning at attention layers and parallel adapters in feedforward layers, achieving superior performance while updating only 6.7\% of parameters \cite{re87_scaling}. Similarly, UniPELT \cite{ref106_unipelt} combines multiple PEFT modules and dynamically selects the most effective components per task, demonstrating strong gains in low-resource scenarios. Both methods significantly outperform individual approaches such as LoRA, Adapters, BitFit, and Prefix Tuning in low-data scenarios. 

In the medical domain, hybrid PEFT strategies hold particular promise due to the inherently complex nature of clinical tasks, which often involve diverse imaging modalities, heterogeneous annotation formats, and domain-specific inductive biases. Hence, hybrid PEFT offers a compelling solution by combining spatially-aware adapters in convolutional blocks with prefix-tuned transformer layers to improve robustness and transferability. Similarly, integrating dynamic prompt tuning to inject contextual text priors with gated adapters for selective cross-modal fusion can improve both semantic alignment and spatial adaptability in multimodal settings. However, such integration introduces new challenges, including increased system complexity and potential optimization conflicts that must be addressed for reliable deployment in clinical workflows \cite{ref68_peft_review}. These issues demand novel solutions, including principled module-selection mechanisms, gradient balancing, and joint training objectives tailored to medical tasks.
\begin{table}[!htbp]
\centering
\begin{threeparttable}
\caption{Comparative summary of major PEFT strategies highlighting trade-offs between efficiency, flexibility, and task performance across different adaptation approaches.}
\label{tab:peft_summary}
\renewcommand{\arraystretch}{1.22}
\setlength{\tabcolsep}{3pt}
\begingroup
\fontsize{9}{11}\selectfont
\begin{tabularx}{\textwidth}{%
  >{\raggedright\arraybackslash\hspace{0pt}}p{2.3cm}    
  >{\raggedright\arraybackslash\hspace{0pt}}p{3.6cm}    
  >{\raggedright\arraybackslash\hspace{0pt}}p{1.5cm}    
  >{\raggedright\arraybackslash\hspace{0pt}}p{2.6cm}    
  >{\raggedright\arraybackslash\hspace{0pt}}X           
}
\toprule
Strategy & Core Mechanism & Params (\%) & Examples & Key Features \\
\midrule
Addition-based & Inject lightweight modules (adapters, prompts, prefixes, side networks) into frozen backbone & \(0.02\text{--}8\%\) & Adapters, Prefix Tuning, Side Tuning; Med-SA, Brain-Adapter & Modular and flexible, supports continual/multimodal learning, sensitive to placement/size, risk of underfitting \\
\midrule
Selection-based & Train only a subset of existing parameters (e.g., biases, sparsity masks) & \(<0.1\%\) & BitFit, DiffPruning, FishMask & Extremely lightweight, effective in low-data settings, mask selection is task-sensitive, risk of instability \\
\midrule
Reparameter-\allowbreak ization-based & Re-express frozen weights with trainable low-rank matrices or structured updates & \(0.01\text{--}0.2\%\) & LoRA, AdaLoRA, DyLoRA, KronA & Strong gains in low-resource regimes, efficient for large ViTs, sensitive to rank choice, uniform adaptation can be suboptimal \\
\midrule
Hybrid & Combine multiple PEFT modules in a single framework & \(1\text{--}7\%\) & MAM Adapter, UniPELT & Leverages complementary strengths, robust in multimodal/complex tasks, higher complexity, risk of optimization conflicts \\
\bottomrule
\end{tabularx}
\begin{tablenotes}
\footnotesize
\item[*] Parameter ranges are approximate and vary with model size, dataset, and task.
\end{tablenotes}
\endgroup
\end{threeparttable}
\end{table}

\subsection{Self-supervised Pretraining on Medical Datasets} \label{ssl_pretraining}
As introduced in Section \ref{ssl}, SSL has emerged as a powerful strategy for pretraining FMs, particularly in the medical domain. Rather than directly fine-tuning FMs (generally trained on natural images) on downstream tasks, an intermediate stage of SSL pretraining on large-scale, unlabeled medical datasets allows the model to acquire domain-relevant representations \cite{ref59_big_ssl}. This step acts as a crucial adaptation phase, improving the model's alignment with medical imaging characteristics before supervised fine-tuning or parameter-efficient adaptation. In this section, we examine the major categories of self-supervised learning methods based on their underlying training strategy. A comparative summary of these categories is provided in Table~\ref{tab:ssl_taxonomy}.

\subsubsection{Traditional Pretext Tasks}
Traditional pretext tasks form the foundation of SSL, using artificial challenges such as predicting geometric transformations (e.g., rotations, flips) or reconstructing jigsaw puzzles to extract generalizable features without requiring manual labels. These tasks are typically hand-crafted and designed around intuitive image manipulations, with the underlying assumption that solving them forces the network to capture useful structural and semantic properties of images \cite{ref1_3d_ssl_methods_medical}. In natural image domains, such strategies have proven effective, with pretrained representations transferring reasonably well to downstream tasks such as classification and detection \cite{ref24_ssl_in_mia}. However, their utility is more constrained in the medical domain. Although they promote attention to structural aspects, they often fall short in capturing the nuanced patterns and subtle diagnostic cues essential for clinical interpretation. For example, simply solving a rotation task or spatial arrangement puzzle may prompt a model to rely on low-level texture or orientation cues, while overlooking rare or context-dependent anomalies that are central to medical diagnostics \cite{ref1_3d_ssl_methods_medical, ref65_transfusion}.

To address these shortcomings, recent research has advanced the field by designing more sophisticated pretext tasks. For instance, Swin-UNETR combines masked inpainting with 3D rotation prediction in a hierarchical transformer framework, enabling the extraction of multi-scale features across entire CT volumes and improving segmentation accuracy in real-world scenarios \cite{ref7_swinunetr}. Taleb et al.\cite{ref109_multimodal_ssl} also extend the traditional jigsaw puzzle task by mixing patches from different imaging modalities within a single puzzle, encouraging the model to learn modality-agnostic features, thereby achieving stronger performance across diverse segmentation benchmarks. Further innovations like the context restoration approach proposed by Chen et al. \cite{ref58_ssl_image_context_restoration}, requires the model to reconstruct meaningful global structure from scrambled or swapped image regions. This not only enables the model to capture both local details and broader context, but also moves beyond the simplicity of basic transformation tasks. Another key advancement is the shift from 2D to 3D pretext tasks, such as 3D rotation prediction or 3D jigsaw puzzles \cite{ref1_3d_ssl_methods_medical}. These volumetric tasks are better aligned with the way medical professionals interpret scans in clinical practice and have demonstrated clear benefits in both effectiveness and accuracy.

Despite their foundational role, traditional pretext tasks exhibit several critical limitations that restrict their effectiveness, particularly in medical imaging. First, their hand-crafted nature introduces task-specific biases, leading models to learn representations tailored to artificial puzzles rather than capturing universally transferable features. As a result, models often overfit to low-level cues such as edges, textures, or patch boundaries, while overlooking subtle but clinically meaningful patterns. Second, many tasks emphasize local structures at the expense of global or contextual understanding, creating a gap between pretraining objectives and downstream requirements such as holistic diagnosis or volumetric segmentation \cite{ref1_3d_ssl_methods_medical, ref58_ssl_image_context_restoration}. Third, the manual design of pretext tasks is labor-intensive and does not scale effectively across different modalities or large transformer-based architectures. These shortcomings have prompted the incorporation of other strategies such as contrastive or hybrid
SSL approaches.

\subsubsection{Contrastive Learning}
Contrastive learning has rapidly emerged as a central paradigm in SSL, enabling models to extract rich and transferable representations from large-scale unlabeled datasets. In this framework, the model learns to align representations of different augmented views of the same image (positive pairs) while pushing apart representations of different images (negative pairs) \cite{ref62_simclr, ref61_survey_contrastive_ssl}. This is typically formalized using a contrastive loss such as InfoNCE or NT-Xent:
\begin{equation}
\mathcal{L}_{\text{contrastive}} = \ell\left(f_{\mathrm{SSL}}(x_i), \; f_{\mathrm{SSL}}(x_j), \; \left\{ f_{\mathrm{SSL}}(x_k) \right\}_{k \in \mathcal{N}} \right),
\end{equation}
where \( x_i \) and \( x_j \) are two augmented views of the same image, \( f_{\mathrm{SSL}} \) is the encoder network, \( \{x_k\}_{k \in \mathcal{N}} \) represents the set of negative samples, and \( \ell(\cdot) \) is a contrastive objective function. Unlike pretext tasks that require explicit manual design, contrastive frameworks learn robust, generalizable features by comparing pairs or groups of data samples. This encourages models to capture intrinsic patterns and structures that remain consistent across variations, enabling the learning of rich, transferable representations directly from large unlabeled datasets. In doing so, contrastive learning directly addresses the task-specific bias problem associated with traditional pretext tasks.

In the medical domain, contrastive learning has shown particular promise, especially where labeled data is scarce but inherently multimodal. Early work like ConVIRT \cite{ref_112_convirt} demonstrated the effectiveness of aligning visual features with paired radiology reports, enabling models to exploit the natural synergy between medical images and textual descriptions through contrastive learning. MICLe \cite{ref59_big_ssl} extends SimCLR \cite{ref62_simclr} by introducing multi-instance contrastive learning, which constructs positive pairs from distinct images of the same patient case to capture intra-class variability. This strategy significantly boosted downstream performance, achieving +6.7\% top-1 accuracy in dermatology and outperforming strong supervised baselines pretrained on ImageNet. While most early approaches focused on 2D or instance-level contrast, more recent work has addressed the unique anatomical context of volumetric medical data. For example, VoCo \cite{ref110_voco} introduced a volume-level contrastive learning framework that distinguishes entire 3D scans as positive or negative pairs, enabling models to learn holistic anatomical representations. When evaluated on six segmentation datasets and one classification task, VoCo consistently outperformed both supervised training and other SSL baselines, achieving improvements of approximately 1–3\% in DSC and classification accuracy. Beyond imaging alone, contrastive objectives have also been incorporated into large-scale FMs for computational pathology and multimodal learning. Vision-language models such as CONCH \cite{ref111_villan_fm} employ dual encoders trained with contrastive loss to align histopathology images and textual descriptions, and have achieved robust zero-shot and cross-modal performance.

Despite its success, contrastive learning faces several key limitations. A central challenge lies in the selection of appropriate positive and negative pairs, as clinically meaningful similarities or differences are often subtle, and careless sampling can lead to false associations or information leakage \cite{ref_112_convirt, ref125_gloria}. Furthermore, contrastive frameworks are highly dependent on effective data augmentations, yet standard strategies may inadvertently distort medically relevant features \cite{ref1_3d_ssl_methods_medical}. Another concern is the reliance on large-scale unlabeled datasets, which can amplify existing biases such as class imbalance, demographic under-representation, and the scarcity of rare conditions. Methodologically, contrastive methods are often restricted to learning global representations, which may overlook fine-grained or spatially localized information that is crucial for clinical tasks. In addition, achieving strong performance typically requires large batch sizes and substantial computational resources \cite{ref62_simclr}, posing scalability challenges, especially for 3D volumetric imaging. Finally, in multimodal contexts such as image–text alignment, ensuring that learned correspondences reflect clinically meaningful semantics rather than superficial associations remains an open problem \cite{ref32_medclip,ref111_villan_fm}. Addressing these challenges may require the use of domain-specific augmentations and the integration of complementary SSL strategies, such as combining contrastive learning with generative or region-level contrastive objectives to capture both coarse and fine-grained semantics.

\subsubsection{Masked Image Modeling}
Masked Image Modeling (MIM) has emerged as a transformative SSL strategy, driven by the success of transformer-based architectures in computer vision. In this approach, the model is tasked with reconstructing randomly masked regions of the input image, compelling it to infer missing information from the visible context. For an image $\mathbf{x}$ and a binary mask $\mathbf{m}$ (where $m_i = 1$ if patch $i$ is masked, $0$ otherwise),
the observed input is $\mathbf{x} \odot (1 - \mathbf{m})$, where $\odot$ denotes element-wise multiplication.
The model is optimized to reconstruct the masked pixels or voxels, typically using a mean squared error (MSE) loss:
\begin{equation}
\mathcal{L}_{\text{MIM}} = \frac{1}{|M|} \sum_{i \in M} \left( x_i - \hat{x}_i \right)^2,
\end{equation}
where $M$ is the set of masked positions, $x_i$ is the ground truth value at position $i$, and $\hat{x}_i$ is the model's prediction for the masked region. This simple yet powerful objective enables the model to learn semantically meaningful representations rather than memorizing low-level cues. By reconstructing masked regions, the model is encouraged to learn semantically meaningful features, capturing fine-grained local details while leveraging broader anatomical context \cite{ref51_mae}. However, standard single-scale MIM may inadequately model long-range dependencies, motivating extensions that explicitly address this limitation \cite{ref154_mim}. Another key advantage of MIM is its reduced dependence on strong data augmentations, which is particularly important in medical imaging where anatomical fidelity must be preserved and artificial distortions can compromise clinical validity.

MIM is particularly appealing in medical imaging, where subtle lesions or abnormalities are often only interpretable within their anatomical context. For example, MAE-based self-pretraining of UNETR improved abdominal CT multi-organ segmentation by 4.7\% DSC and enhanced brain tumor segmentation by 1.5\% DSC, outperforming other U-Net architectures \cite{ref26_ssl_mae_mia}. Similarly, MAE and SimMIM \cite{ref114_simmim} have been adopted to volumetric CT and MRI, enabling ViT backbones to learn robust, context-aware representations achieving 4–5\% higher DSC on multi-organ CT segmentation compared to contrastive approaches, while also accelerating supervised fine-tuning by up to 1.4× \cite{ref52_mim3d}. Mask-in-Mask (MiM) \cite{ref154_mim} introduces multi-level reconstruction and cross-level alignment to overcome the limitations of single-scale MAE, such as limited global context and weak multi-scale consistency. By reconstructing multi-level masked volumes, MiM captures both fine-grained anatomical details and global structure, achieving 85.4\% DSC on BTCV, 70.1\% DSC on MSD, and 94.3\% ACC and 99.7\% AUC on CC-CCII, outperforming prior SSL methods and showing strong cross-modality transfer to MRI (BraTS21, 79.9\% DSC).

Despite its progress, MIM faces critical challenges in the medical domain. The effectiveness of MIM depends strongly on the choice of masking ratio, shape, and distribution. Suboptimal designs can lead to either trivial reconstructions or excessive information loss, thereby degrading the quality of learned representations \cite{ref51_mae, ref26_ssl_mae_mia}. Moreover, standard random masking strategies often overlook subtle yet clinically significant variations, limiting the model’s ability to capture pathological features \cite{ref113_mmclip}. This issue is particularly relevant for existing methods such as MAE and SimMIM, which primarily focus on pixel- or voxel-level reconstruction. While effective in general-purpose vision tasks, these approaches may struggle to learn clinically meaningful features in ambiguous or anatomically variable regions \cite{ref25_ssl_swin_trans}. To address these issues, more advanced models like MMCLIP \cite{ref113_mmclip} move beyond random masking to design a cross-modal, attention-guided masking approach tailored for medical images. Additionally, incorporating radiomic priors (quantitative feature descriptors), and combining MIM with contrastive or multimodal objectives represent a promising shift toward more clinically meaningful and effective SSL pretraining strategies.

\subsubsection{Hybrid SSL}
Hybrid approach strategically integrates multiple SSL paradigms and has emerged as a promising solution to the limitations of single-objective approaches. The motivation for hybrid approaches stems from the recognition that traditional pretext tasks encourage the model to learn low and mid-level semantic and spatial relationships, contrastive methods promote invariance and global discrimination, while MIM enable the model to capture fine-grained and context-sensitive structures \cite{ref30_contrastive_learning,ref26_ssl_mae_mia}. However, when applied independently, these strategies may lead to incomplete or biased representations in complex medical settings. For instance, Contrastive Masked Autoencoders (CMAE) \cite{ref115_cmae} combine MIM and contrastive learning by simultaneously reconstructing masked patches to retain local context, while enforcing global instance-level discrimination through contrastive loss. This dual objective leads to more robust and transferable representations, achieving 85.3\% top-1 accuracy on ImageNet. Such empirical evidence from computer vision highlights the potential of hybrid SSL as a promising avenue for MIA.

\begin{table}[htb!]
\centering
\begin{threeparttable}
\caption{Comparative summary of hybrid SSL methods in medical imaging, illustrating integrated objectives, modalities, and key trade-offs in performance.}
\label{tab:hybrid_ssl_summary}
\renewcommand{\arraystretch}{1.2}
\setlength{\tabcolsep}{3pt}
\begingroup
\fontsize{9}{11}\selectfont
\begin{tabularx}{\textwidth}{%
    >{\raggedright\arraybackslash}p{2.3cm}   
    >{\raggedright\arraybackslash}p{3.5cm}   
    >{\raggedright\arraybackslash}p{2.6cm}   
    >{\raggedright\arraybackslash}X          
    >{\raggedright\arraybackslash}X          
}
\toprule
Method & Hybrid Objectives & Modality & Key Innovations & Findings / Limitations \\
\midrule
HybridMIM \cite{ref116_hmim} & Pixel/region-level MIM + sample-level contrastive & 3D CT/MRI & Two-level masking, dynamic region selection, SimCSE contrastive & SOTA segmentation, faster pretraining, no inference acceleration \\
\midrule
MMCLIP \cite{ref113_mmclip} & Contrastive (image–text) + MIM + MLM & 2D chest X-ray + paired/unpaired text & Cross-modal attention-masked modeling, entity-driven MLM, prompt-driven attention & SOTA zero-shot, competitive with MedKLIP/GLORIA, limited by NER accuracy and data size \\
\midrule
CONCH \cite{ref111_villan_fm} & Image–text contrastive + captioning loss + unimodal SSL (iBOT, autoregressive LM) & 2D histopathology (WSI tiles) + text (captions/reports) & Large-scale curation of 1.17M image–caption pairs, CoCa-style hybrid pretraining & SOTA in zero-/few-shot classification, segmentation, retrieval, captioning, limited in rare/open-set classes \\
\midrule
SwinUNETR \cite{ref25_ssl_swin_trans} & Masked image inpainting + 3D rotation prediction + patch-level contrastive & 3D CT & Multi-task pretraining on 5,050 CTs, hierarchical transformer encoder, joint multi-loss & SOTA on BTCV/MSD segmentation, boosts label
efficiency (+10\% DSC with 10\% labels) \\
\bottomrule
\end{tabularx}
\begin{tablenotes}
\footnotesize
\item[*] Reported results are from respective publications, benchmarks, datasets, and backbones may differ.
\end{tablenotes}
\endgroup
\end{threeparttable}
\end{table}

In medical imaging, various studies have explored different hybrid SSL strategies, as summarized in Table \ref{tab:hybrid_ssl_summary}. Beyond existing approaches, integrating MIM with contrastive learning could yield richer representations by capturing complementary aspects of the image, fine-grained, context-sensitive structures from MIM and global discrimination from contrastive learning. Likewise, combining masked image modeling with radiomics feature prediction can guide the model not only to reconstruct missing regions but also to encode quantitative descriptors such as texture, shape, and intensity statistics, thereby aligning learned representations more closely with clinically meaningful biomarkers. While hybrid SSL pretraining holds great promise, its implementation presents significant challenges. First, balancing multiple SSL objectives is a complex task, as improper weighting can lead to task interference or under-utilization of certain pretext signals, ultimately degrading learned representations. Second, designing effective hybrid architectures, particularly for multi-modal data requires careful handling of modality-specific features and cross-modal interactions, as naive fusion can dilute clinically relevant information \cite{ref32_medclip}. Third, increased model and training complexity can result in higher computational costs, unstable convergence, and difficulties in hyperparameter tuning. Additionally, there is a risk of overfitting to synthetic pretext tasks or domain artifacts if hybrid objectives are not well aligned with clinical goals \cite{ref116_hmim}. Finally, rigorous evaluation on clinically meaningful downstream tasks is essential, as apparent gains on proxy metrics may not translate to real-world utility \cite{ref111_villan_fm}.

\begin{table}[htb!]
    \centering
    \renewcommand{\arraystretch}{1.2}
    \setlength{\tabcolsep}{3pt}
    \caption{Comparative summary of SSL strategies in medical imaging, highlighting differences in core mechanisms, representative methods, strengths, and limitations.}
    \label{tab:ssl_taxonomy}
    \begingroup
    \fontsize{9}{11}\selectfont
    \begin{tabularx}{\textwidth}{%
        >{\raggedright\arraybackslash}p{2.0cm}
        >{\raggedright\arraybackslash}p{3.0cm}  
        >{\raggedright\arraybackslash}p{3.2cm}  
        >{\raggedright\arraybackslash}X          
        >{\raggedright\arraybackslash}X          
    }
        \toprule
        Category & Core Mechanism & Representative Methods & Strengths & Limitations \\
        \midrule
        Traditional Pretext & Handcrafted auxiliary tasks (rotation, jigsaw, inpainting, context restoration) & Inpainting \cite{ref7_swinunetr}, 3D Jigsaw, 3D Rotation \cite{ref1_3d_ssl_methods_medical}, Context Restoration \cite{ref58_ssl_image_context_restoration}, Multimodal Jigsaw \cite{ref109_multimodal_ssl} & Simple, interpretable, low compute, captures structural cues & Limited semantic depth, struggles with subtle features, scaling to 3D is costly \\
        \midrule
        Contrastive Learning & Align positive pairs while separating negatives (InfoNCE-style) & ConVIRT \cite{ref_112_convirt}, MICLe \cite{ref59_big_ssl}, VoCo \cite{ref110_voco}, CONCH \cite{ref111_villan_fm}, MedCLIP \cite{ref32_medclip} & Learns transferable invariances, effective in multimodal/cross-domain tasks & Sensitive to sampling, needs large datasets, limited fine-grained feature capture \\
        \midrule
        Masked Image Modeling & Mask patches/voxels and reconstruct for local + global context & SelfMedMAE \cite{ref26_ssl_mae_mia}, SimMIM, MAE \cite{ref52_mim3d}, MMCLIP \cite{ref113_mmclip}, MiM \cite{ref154_mim} & Captures anatomical context, strong transferability, less dependent on augmentations & Highly dependent on mask design, may miss subtle pathology, expensive for 3D \\
        \midrule
        Hybrid SSL & Combine multiple paradigms (contrastive + generative, etc.) & CMAE \cite{ref115_cmae}, HybridMIM \cite{ref116_hmim}, MMCLIP \cite{ref113_mmclip}, Swin-UNETR \cite{ref7_swinunetr}, CONCH \cite{ref111_villan_fm} & Balances complementary strengths, robust and transferable representations & Higher training complexity, balancing objectives can cause conflicts, more compute required \\
        \bottomrule
    \end{tabularx}
    \endgroup
\end{table}

\subsection{Multimodal and Cross-modal Adaptations}
Unlike traditional approaches, multimodal and cross-modal adaptation leverages multiple data sources (typically images, videos, texts) to learn clinically meaningful representations. In medical imaging, text-only supervision (e.g., labels extracted from radiology reports) often inherits reporting bias, hedging, and omission, and may misalign with the image. On the other hand, image-only models lack clinical context, making it difficult to distinguish visually similar patterns such as postoperative changes and tumor recurrence. By combining visual data with contextual information such as patient history, prior studies, and protocol metadata, multimodal models can learn richer, more robust representations that generalize better across settings \cite{ref32_medclip, ref105_medblip}. This holistic approach is particularly valuable in medical domain where diagnostic reasoning often integrates visual and textual evidence, and lacks quality annotated data \cite{ref29_multimodal_report_generation,ref50_transmed, ref109_multimodal_ssl}. In doing so, multimodal adaptation not only compensates for label scarcity but also improves model generalizability, robustness, and clinical applicability across diverse tasks and clinical environments. 

Recent advances in vision-language models, inspired by breakthroughs like CLIP and BLIP have inspired new research directions in the medical domain. For instance, MediCLIP \cite{ref146_mediclip} adapts CLIP for medical anomaly detection by combining learnable prompts, adapter-based vision–text alignment, and synthetic anomaly generation, achieving SOTA few-shot detection and localization on CheXpert, BrainMRI, and BUSI with $\sim10\%$ higher AUROC than existing methods and strong zero-shot generalization across tasks. ConVIRT leverages SSL to learn rich representation from paired chest X-rays and reports achieves comparable performance with roughly 10\% of the labeled data compared to ImageNet initialized baselines \cite{ref_112_convirt}. GLoRIA leverages multimodal image–report alignment to achieve strong label efficiency, reaching AUROC of 86.6 on CheXpert and 86.1 on RSNA with only 1\% of labels, outperforming ImageNet-pretrained models trained with 100\% labels \cite{ref125_gloria}. Multimodal adaptations like BioCLIP, MedBLIP and BioMedCLIP extend these foundations by curating large-scale, domain-specific image–text datasets and employing contrastive or cross-modal pretraining for joint visual-textual alignment. These models set new benchmarks in zero-shot classification, image-report retrieval, and report generation, often outperforming generic models on tasks like chest X-ray interpretation and clinical question answering \cite{ref124_Biodmedclip,ref105_medblip,ref28_bioclip}. 

Cross-modal adaptation extends the benefits of multimodal learning to settings where only one modality is available or labeled data is scarce. 
For instance,  Tiu et al. \cite{ref155_expert} align chest X-rays with reports to enable zero-shot multi-label classification, achieving radiologist-level performance on CheXpert (F1 = 0.606 vs. 0.619) while generalizing robustly to external datasets, despite using no labeled samples. Similarly, Cao et al. \cite{ref156_bootstrapping} distill language-guided knowledge from 2D chest X-ray experts into 3D CT models through report driven retrieval and dual distillation, attaining superior performance across cross-modal tasks, including zero-shot CT diagnosis (F1 = 33.0 on ChestCT-16K and 65.8 on external LIDC nodules) and report generation (F1 = 37.3 vs.\ $\leq$ 35.4 for BLIP). This enables efficient image-only inference, reaching performance levels comparable to radiologists on specific lung pathologies. Finally, MedCLIP demonstrates that training on unpaired images and texts can effectively scale cross-modal supervision while preserving strong zero-shot transfer across medical imaging tasks \cite{ref32_medclip}.

However, these advances face persistent and domain-specific challenges. High-quality image–report pairs remain limited, often tied to single institutions, and reporting practices can introduce bias that undermines generalizability \cite{ref29_multimodal_report_generation}. A substantial semantic gap also exists between medical and natural domains, as anatomical structures and clinical concepts do not align well with representations learned from generic image–text corpora \cite{ref65_transfusion, ref3_interobserver_variability, ref9_fm_medical_comprehensive_survey}. Furthermore, report generation and visual question answering models risk hallucinating findings, misrepresenting uncertainties, or perpetuating existing reporting biases, posing serious risks for clinical deployment unless thoroughly validated \cite{ref33_comprehensive_survey_fm, ref10_fm}. The use of real clinical text further introduces privacy and governance obligations, while the training and deployment of large multimodal models requires significant computational resources, with implications for accessibility, environmental sustainability, and regulatory compliance. Overcoming these barriers demands evaluation protocols that move beyond accuracy to include factuality, calibration, selective prediction, and robustness across sites and reporting styles, alongside deployment strategies that safeguard privacy and resource constraints.

%% file: peft_taxonomy.tex
\begin{figure}[ht]
    \centering
    \resizebox{\textwidth}{!}{%
    \begin{tikzpicture}[
        >=Stealth, 
        font=\sffamily, 
        every node/.style={rounded corners=5pt, draw=blue!50, fill=white, anchor=west, align=left},
        branch/.style={fill=blue!7, draw=blue!50, font=\normalsize, text width=3.7cm, align=left},
        edge from parent/.style={draw, thick, -{Stealth[scale=1.0]}, line width=1.1pt},
        level distance=2.8cm, 
        sibling distance=0.9cm,
        inner sep=6pt
    ]

\node[rectangle, draw=blue!50, fill=white, font=\large\bfseries, rotate=90, anchor=center, inner sep=6pt] 
(root) at (-1.0,-5.0) {PEFT Methods for MIA};

\draw [line width=1.1pt] (root.south) -- ++(0.65,0);
\draw[line width=1.1pt] (0.02,-0.3) -- (0.02,-8.6);

\node[branch] (add)    at (0.5,-0.3) {Addition-based Method (§\ref{adition-based})};
\node[branch] (select) at (0.5,-5.0) {Selection-based Method (§\ref{selection-based})};
\node[branch] (repara) at (0.5,-6.8) {Reparameterization-Based Method (§\ref{reparameterization-based})};
\node[branch] (hyb)    at (0.5,-8.6) {Hybrid Method (§\ref{hybrid-methods})};

\draw[line width=1.1pt] (0,-0.3) -- (add.west);
\draw[line width=1.1pt] (0,-5.0) -- (select.west);
\draw[line width=1.1pt] (0,-6.8) -- (repara.west);
\draw[line width=1.1pt] (0,-8.6) -- (hyb.west);

\draw [line width=1.1pt] (add.east) -- ++(1.2,0) coordinate (addbranch);

\node[fill=white, draw=blue!50, text width=2.6cm, right=2.0cm of add.east, yshift=2.0cm, anchor=west] (adapter) {Adapter Tuning };
\node[fill=white, draw=blue!50, text width=2.6cm, right=2.0cm of add.east, yshift=0cm, anchor=west] (prompt) {Prompt Tuning };
\node[fill=white, draw=blue!50, text width=2.6cm, right=2.0cm of add.east, yshift=-1.8cm, anchor=west] (prefix) {Prefix Tuning };
\node[fill=white, draw=blue!50, text width=2.6cm, right=2.0cm of add.east, yshift=-3.4cm, anchor=west] (side) {Side Tuning};

\node[fill=white, draw=blue!50, text width=10cm, right=2.0cm of adapter.east, anchor=west] (adapter1) {\textbf{Segmentation:} Med-SA \cite{ref53_medsa}, MA-SAM \cite{ref127_masam},Trans-SAM \cite{ref129_trans-sam}, 3DSAM-adapter \cite{ref130_3dsam_adapter},  \textbf{Classification: }Brain-Adapter \cite{ref86_brain_adapter}, TSA \cite{ref96_peft_mia} };

\node[fill=white, draw=blue!50, text width=10cm, right=2.0cm of prompt.east, anchor=west] (prompt1) {\textbf{Segmentation:} PUNETR \cite{ref88_prompt_tuning1}, AutoSAM \cite{ref55_autosam}, DVPT \cite{ref89_prompt_tuning} \textbf{Classification:} DVPT \cite{ref89_prompt_tuning}, EPT \cite{ref90_prompt_tuning}, KBDPT \cite{ref131_kbdpt}};

\node[fill=white, draw=blue!50, text width=10cm, right=2.0cm of prefix.east, anchor=west] (prefix1) {\textbf{Radiology Report:} RadAdapt \cite{ref141_radadapt}, \textbf{Visual Question Answering:}  \cite{ref132_prefix}};

\node[fill=white, draw=blue!50, text width=10cm, right=2.0cm of side.east, anchor=west] (side1) {\textbf{Segmentation: }Ladder ST \cite{re133f_ladder_st}, \textbf{Classification:} Fushion ST \cite{ref134_fushion_st}};

\draw[-, line width=1.1pt] (adapter.east) -- (adapter1.west);
\draw[-, line width=1.1pt] (prompt.east) -- (prompt1.west);
\draw[-, line width=1.1pt] (prefix.east) -- (prefix1.west);
\draw[-, line width=1.1pt] (side.east) -- (side1.west);

\draw[-, line width=1.1pt] (addbranch |- adapter.west) -- (adapter.west);
\draw[-, line width=1.1pt] (addbranch |- prompt.west) -- (prompt.west);
\draw[-, line width=1.1pt] (addbranch |- prefix.west) -- (prefix.west);
\draw[-, line width=1.1pt] (addbranch |- side.west) -- (side.west);

\draw [line width=1.1pt] (addbranch |- adapter.west) -- (addbranch |- side.west);

\draw [line width=1.1pt] (select.east) -- ++(1.2,0) coordinate (selectbranch);

\node[fill=white, draw=blue!50, text width=15cm, right=2.0cm of select.east, anchor=west] (bitfit) {\textbf{Classification: }BitFit \& Bias Tuning \cite{ref96_peft_mia} };

\draw[-, line width=1.1pt] (selectbranch |- bitfit.west) -- (bitfit.west);

\draw [line width=1.1pt] (repara.east) -- ++(1.2,0) coordinate (reparabranch);

\node[fill=white, draw=blue!50, text width=15cm, right=2.0cm of repara.east, anchor=west] (reparaNode) {\textbf{Segmentation: }SAMed \cite{ref54_SAMed}, AFTer-SAM \cite{ref135_aftersam}, BC-SAM \cite{ref138_bcsam} \textbf{Classfication: }MedBLIP
\cite{ref105_medblip}, MeLo \cite{ref136_melo}, LoRA \cite{ref96_peft_mia, ref137_lian2024less}, BC-SAM \cite{ref138_bcsam}, LoRA \cite{ref140} \textbf{Visual Question Answering:} MedBLIP, \cite{ref105_medblip}, \textbf{Report Generation: } \cite{ref141_radadapt}, LoRA \cite{ref139_report_generatio}, \textbf{Lesion Detection: }LoRA \cite{ref140}};

\draw[-, line width=1.1pt] (reparabranch |- reparaNode.west) -- (reparaNode.west);

\draw [line width=1.1pt] (hyb.east) -- ++(1.2,0) coordinate (hybbranch);

\node[fill=white, draw=blue!50, text width=15cm, right=2.0cm of hyb.east, anchor=west] (hybNode) {\textbf{Segmentation:} Med-SA \cite{ref53_medsa}, \textbf{Classification:} \cite{ref162_probing,ref164,ref163_sparsity}};

\draw[-, line width=1.1pt] (hybbranch |- hybNode.west) -- (hybNode.west);

\end{tikzpicture}
    }
    \caption{Taxonomy of PEFT methods for MIA, highlighting their adaptation and use across a range of medical tasks.}\label{fig:peft-taxonomy}
\end{figure}
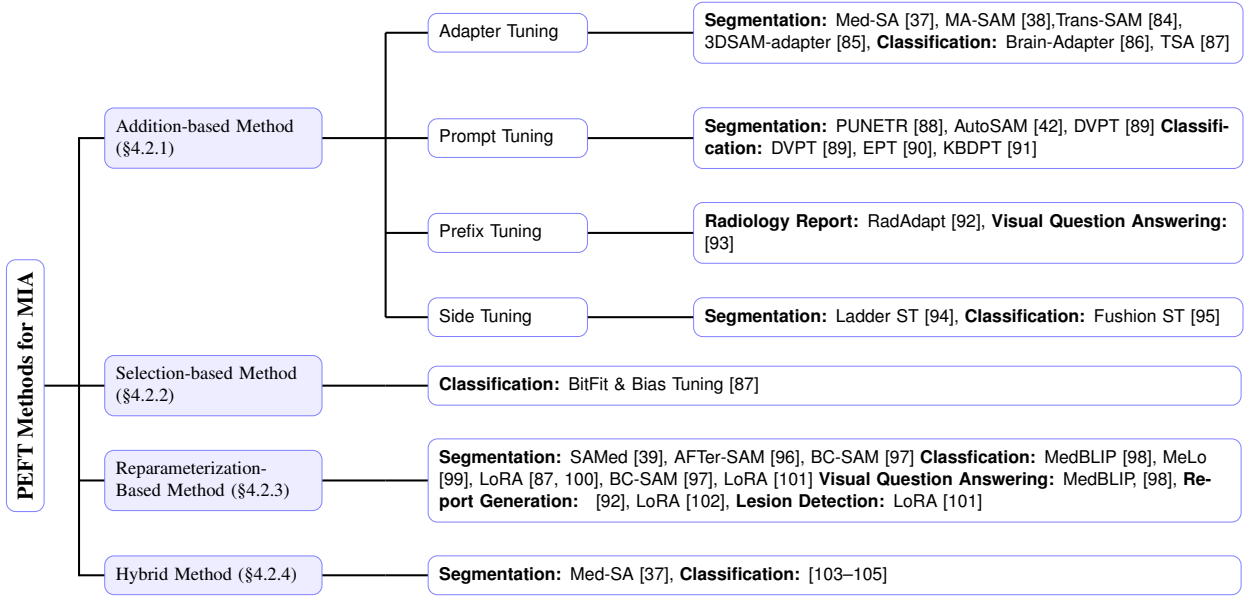

%% file: 05_applications.tex
\section{Applications of Foundation Models in Medical Image Analysis}\label{sec5:applications}
FMs are transforming MIA by introducing generalizable, data-efficient solutions that excel across diverse clinical settings. This section highlights recent advances and applications of FMs in core areas of medical imaging tasks. A summary of representative methods and adaptation strategies is provided in Table~\ref{tab:fm_adapt_summary}.

\subsection{Detection or Localization}
Accurate detection and localization of abnormalities remain among the most critical yet persistently challenging tasks in MIA. Unlike segmentation, which benefits from spatial continuity and well-defined region-based supervision, detection must identify often tiny, ambiguous, or low-contrast abnormalities embedded within complex anatomical backgrounds. These challenges are amplified by extreme class imbalance, where rare but clinically significant findings must be distinguished from abundant normal tissue. Moreover, variability in imaging protocols further complicates generalization, while subtle contextual cues, such as anatomical relationships or temporal changes may be crucial for accurate localization but are difficult for models to capture \cite{ref43_survey_dl_mia}. Recent innovations have begun to address these challenges by combining large-scale pretraining, self-supervised learning, and modular FM integration. For example, MedLSAM \cite{ref122_medlsam} automates 3D localization and segmentation by integrating a self-supervised localization module trained on 14,000 CT scans with SAM. This approach reduces slice-level annotation to a few extreme-point prompts while achieving cross-organ DSC above 69\%, thereby overcoming annotation bottlenecks, poor cross-organ generalization, lack of 3D priors, and excessive manual interaction. BiomedParse \cite{ref123_BiomedParse} jointly performs segmentation, detection, and recognition across nine imaging modalities using 6 million image–mask–text triples, achieving new SOTA performance with a median DSC of 0.942 in pathology cell segmentation and  0.857 (39.6\% higher) DSC on irregular-shaped objects. By eliminating bounding-box dependence, BiomedParse solves key challenges of scalability, irregular morphology, and semantic inconsistency across datasets. However, it still relies on harmonizing massive image–mask–text datasets across modalities and requires significant computational resources. In digital pathology, the Virchow FM \cite{ref143_virchow} enables SOTA pan-cancer and rare cancer detection from whole-slide images (WSIs), achieving an AUC of 0.95 across diverse cancer types. Leveraging over 1.5 million WSIs and self-supervised learning, Virchow generalizes well to rare and out-of-distribution cases, matching or exceeding specialist models while requiring significantly fewer annotations, highlighting the promise of FMs for robust detection in low-data and rare-disease settings.

To address persistent challenges in detection and localization, several innovative strategies are being explored. FM-based anatomical priors, when combined with large-scale self-supervised pretraining offer universal structural context that reduces reliance on exhaustive bounding box or voxel annotations. Prompt-based methods, such as extreme-point or bounding-box guidance, further reduce annotation burden while enabling precise target identification \cite{ref21_sam, ref122_medlsam}. Safety and reliability can be strengthened through uncertainty-aware approaches, including Bayesian inference and ensemble modeling, which highlight ambiguous predictions for expert review. Robustness is also improved by multimodal integration, where imaging signals are complemented with clinical reports, genomic data, or spatial transcriptomics to overcome noise and variability \cite{ref33_comprehensive_survey_fm,ref35_fm_advancing_healthcare}. Moreover, interactive frameworks that allow clinicians to rapidly refine predictions promote both efficiency and trust in clinical practice. Collectively, these directions signal a transition from narrow, fully supervised detectors toward generalizable and clinically aligned localization systems capable of scaling across organs, modalities, and diverse healthcare environments.

\subsection{Segmentation}
Segmentation has emerged as the flagship application of FMs in MIA, driving most high-impact advances in recent years. Unlike classification or detection, segmentation provides precise, clinically actionable information, such as the exact location, boundaries, and volume of anatomical structures or lesions that are critical for clinical assessment. Additionally, the introduction of powerful, general-purpose segmentation models like SAM has accelerated interest in this space, offering a flexible, prompt-driven interface that supports a wide range of segmentation tasks across domains \cite{ref41_sam_for_miseg,ref37_fm_misegmentation}. Yet, directly adopting SAM to medical images often yields suboptimal performance due to inherent domain-specific challenges and complex anatomical structures unique to medical data \cite{ref31_medsam}. 

To address this gap, recent efforts have shifted toward domain-adapted segmentation FMs by training on large, diverse, and annotated medical datasets. For example, MedSAM and MedSAM2 \cite{ref157_medsam2} retain SAM’s flexible prompting interface, but adapt SAM to the medical domain by training on large, modality-rich medical datasets (1.5M images and 455K 3D image–mask pairs with 76K video frames, respectively). Empirical evaluations show that MedSAM consistently outperforms both SAM and modality-specific models, achieving improvements up to 10\% in DSC on external validation tasks \cite{ref31_medsam}. Likewise, MedSAM2 pushes these advances further by enabling segmentation in both 3D image and video modalities, achieving median DSC of 88.8\% on CT, 88.4\% on MRI, and 87.2\% on PET. Beyond accuracy, its human-in-the-loop pipeline reduced average lesion annotation time from over 500 seconds to just 65–74 seconds per case in CT and MRI, and a 92\% reduction in cardiac ultrasound video annotation time, thereby reducing overall annotation costs by more than 85\% \cite{ref157_medsam2}. Similarly, SAM-Med3D \cite{ref119_sammed3d} trained from scratch on 22K 3D volumes and 143K masks spanning 245 categories, achieved a 60.1\% relative DSC improvement over SAM with only one prompt per volume, while cutting inference time to as little as 1–26\% of SAM’s runtime, offering both higher accuracy and clinical efficiency. Extending these advances, VISTA3D \cite{ref121_vista3d} addresses core challenges in 3D medical segmentation by replacing inefficient slice-wise annotation with a unified 3D backbone that delivers accurate automatic segmentation. It also introduces an interactive correction branch that allows clinicians to refine results with minimal effort, and a supervoxel distillation strategy to transfer semantic knowledge from 2D FMs into 3D.
These features enables robust zero-shot generalization to unseen structures and reduce the gap between expert-trained models and practical clinical use \cite{ref121_vista3d}. SegVol \cite{ref120_segvol} complements this by emphasizing universality and efficiency, addressing the fragmentation of task-specific models by training across diverse datasets and anatomical categories ( 90K unlabeled and 6K labeled CTs spanning 200+ categories), and introducing a zoom-out–zoom-in mechanism that drastically reduces the computational cost of volumetric inference while preserving precision. By combining semantic and spatial prompts, it also resolves the ambiguity of interactive segmentation, ensuring reliable results across both organs and lesions \cite{ref120_segvol}. Together, they demonstrate that clinically deployable 3D FMs are both adaptable to new domains and robust across diverse anatomies.

Despite rapid progress, medical image segmentation often requires delineating subtle, irregular, low contrast, or ambiguous boundaries. The shapes, sizes and boundaries of the lesion can vary dramatically between patients and disease states, requiring models to excel in both precision and flexibility \cite{ref31_medsam,ref120_segvol}. These challenges are amplified in volumetric segmentation, where moving from 2D to 3D requires handling complex spatial dependencies and substantially higher computational costs. At the same time, clinical applications demand near-zero tolerance for false positives or boundary errors, as segmentation inaccuracies can directly influence diagnosis and treatment planning.  As a result, future progress hinges on embedding domain-specific priors and anatomical constraints to preserve structural plausibility, as well as boundary or structure-aware objectives that explicitly preserve continuity and fine morphological details. Moreover, rigorous validation on clinically relevant metrics such as tumor burden, surgical margin error, or treatment response consistency must guide model optimization, ensuring that segmentation outputs move beyond pixel overlap toward trustworthy, outcome-aligned clinical decision support \cite{ref_40_sammed2d}.

\subsection{Classification}
FMs are redefining medical image classification by extending multimodal pretraining to align visual representations with rich clinical text. Pioneering approaches such as MedCLIP \cite{ref32_medclip}, GLoRIA \cite{ref125_gloria}, and BiomedCLIP \cite{ref124_Biodmedclip} adapt general vision–language architectures to the medical domain, using contrastive learning on paired or weakly paired data to capture subtle diagnostic cues while reducing dependence on costly manual labels. BiomedCLIP, in particular, scales this approach by incorporating domain-specific language encoders and vast biomedical data, resulting in strong zero-shot and few-shot performance across diverse modalities \cite{ref124_Biodmedclip}. At the same time, novel frameworks are beginning to address fundamental barriers to equitable clinical deployment. For example, TFA-LT mitigates long-tailed distributions through lightweight linear adapters and ensemble strategies, reporting up to 27.1\% accuracy gains on rare classes with only a fraction of the compute required by prior methods \cite{ref126_tfa-lt}. In computational pathology, the UNI model, pretrained on 100M+ histopathology tiles, demonstrates unprecedented scalability, achieving robust cross-tissue generalization and accurate disease/tissue classification even under diagnostically ambiguous conditions \cite{ref144_Uni}.

A promising new direction lies in solutions that move beyond conventional supervised pipelines toward systems that are inherently adaptive, trustworthy, and clinically aligned. Report-guided hybrid supervision leverages the abundance of radiology text to generate weak yet informative labels, substantially reducing dependence on costly manual annotation. Beyond label efficiency, anatomy- and physics-grounded causal priors constrain models to focus on clinically relevant structures while ignoring superficial cues, improving generalization under real-world conditions. At the deployment stage, selective and fairness-aware classification reframes prediction as a risk-controlled process, ensuring that model outputs remain reliable under uncertainty and that calibration quality is maintained equitably across patient subgroups \cite{ref34_generalist_fm, ref36_vfm_mia}. Robustness under domain shift can be further strengthened through parameter-efficient domain adaptation and test-time learning, where lightweight adapters continually adjust to new scanners or sites without retraining the full backbone \cite{ref84_unified_peft,ref96_peft_mia}. Finally, cost-sensitive evaluation frameworks prioritize metrics tied to patient outcomes, ensuring that model optimization reflects the asymmetric risks of false positives and false negatives in clinical workflows. Together, these approaches shift the field toward classifiers that are not static predictors but dynamic, context-aware decision partners capable of delivering equitable, transparent, and clinically meaningful outcomes.

\begin{table}[!htbp]
\centering
\begin{threeparttable}
\caption{Comparative summary of FM adaptation strategies in medical imaging, highlighting adaptation mechanisms, modalities, and trade-offs.}
\label{tab:fm_adapt_summary}
\renewcommand{\arraystretch}{1.08}
\setlength{\tabcolsep}{2pt}
\begingroup
\fontsize{8.0}{9}\selectfont
\sloppy 
\begin{tabularx}{\textwidth}{%
    >{\RaggedRight\arraybackslash}p{2.0cm}   
    >{\RaggedRight\arraybackslash}p{2.3cm}   
    >{\RaggedRight\arraybackslash}p{2.1cm}   
    >{\RaggedRight\arraybackslash}X          
    >{\RaggedRight\arraybackslash}X          
}
\toprule
Method & Task & Modality & Adaptation & Findings / Limitations \\
\midrule
Med-SA \cite{ref53_medsa} & Interactive and automatic segmentation & CT, MRI, US, Fundus, Dermoscopy & Adapters in encoder/decoder, SD-Trans for 3D, HyP-Adpt for prompt conditioning, only 2\% SAM params updated & SOTA on 17 tasks, BTCV Dice $\uparrow$2.9\% vs.\ Swin-UNetr, $20\times$ fewer params than MedSAM, robust generalization but prompt ambiguity remains \\
\midrule
SAMed \cite{ref54_SAMed} & Multi-organ segmentation & CT (Synapse) & LoRA on encoder, prompt encoder, mask decoder ($\sim$5\% params) & Dice 81.9, HD 20.6, 6.3M params for compact version, prompt-free inference, below SOTA, lacks 3D spatial context \\
\midrule
AutoSAM \cite{ref55_autosam} & Semantic segmentation & Histopathology & Train prompt encoder to generate embeddings, SAM frozen, no FM fine-tuning & SOTA Dice/IoU: MoNuSeg 82.4/70.2, GlaS 92.8/87.1, strong on polyp datasets, cross-domain generalization not yet tested \\
\midrule
MA-SAM \cite{ref127_masam} & Automatic and prompt-based segmentation & CT, MRI, Surgical video & FacT 3D adapters in each transformer block, capture volumetric and temporal info, prompts optional & Dice $\uparrow$0.9\% (CT multi-organ), $\uparrow$2.6\% (MRI prostate), $\uparrow$9.9\% (surgical scene), prompts $\uparrow$38.7\% Dice, higher complexity from 3D adapters \\
\midrule
MVFA-AD \cite{ref145_mvfa} & Anomaly classification and segmentation (zero/few-shot) & CT, MRI, X-ray, Retinal OCT, Histopathology & Residual adapters in CLIP, pixel-wise alignment, dual-branch comparison, prompt cues for anomaly localization & Outperforms SOTA on BMAD, AUC $\uparrow$6.24\% (classification), $\uparrow$2.03\% (segmentation) zero-shot, strong cross-domain transfer, only adapters updated \\
\midrule
MediCLIP \cite{ref146_mediclip} & Few-shot anomaly detection and localization & CXR, Brain MRI, Breast US & Adapter-based PEFT on CLIP, learnable prompts, synthetic anomaly generation for contrastive training & SOTA on CheXpert, Brain MRI, BUSI, $\sim$10\% gain with $<1$\% data, 94\% full-shot with 1\% images, robust zero-shot, dependent on anomaly realism \\
\midrule
MeLo \cite{ref136_melo} & Disease classification & CXR, Mammography, Blood smear & LoRA modules in ViT self-attention layers, backbone frozen, parameter-efficient fine-tuning (0.14--0.17\% params) & Matches or exceeds full FT, AUC improves with model size, efficient multi-task deployment, rapid model switching, requires high-quality pretraining \\
\midrule
PromptMRG \cite{ref147_promptmrg} & Medical report generation & Chest X-ray (MIMIC-CXR, IU X-Ray) & Retrieval-augmented CLIP features, prompt-based decoding, retrieval aids classification & SOTA Clinical Efficacy (F1) on both datasets, better rare disease recognition, retrieval improves classification more than generation \\
\bottomrule
\end{tabularx}
\endgroup
\begin{tablenotes}
\footnotesize
\item[*] Summarized from respective publications, benchmarks and datasets differ across studies.
\end{tablenotes}
\end{threeparttable}
\end{table}

%% file: 06_trends.tex
\section{Trends and Future Directions} \label{sec6:trends}
Building on our critical review, we identify five converging trends that will define the next era of FM adaptation in MIA. For each, we highlight a core limitation in the literature and propose transformative research directions to build adaptive, trustworthy, and clinically integrated FMs that reflect the realities of healthcare practice.
\newline
\subsection{From Static Deployment to Continual Adaptation}
 A major limitation of current FM deployment in medical imaging is their static nature and once adapted to a target clinical task, models are rarely updated to reflect evolving data distributions, imaging protocols, or emergent clinical needs. This static approach is increasingly misaligned with real-world healthcare, where medical data and practices continuously evolve. Future research must prioritize continual learning and lifelong adaptation frameworks that enable FMs to ingest new data and refine their representations while preserving previously acquired knowledge. Such advancements will require innovations in memory-efficient fine-tuning, such as rehearsal-based learning, replay buffers, and regularization-based consolidation \cite{ref148_experience_replay}, alongside robust online domain adaptation mechanisms that respond to distribution shifts in real time. For example, the introduction of new imaging scanners or changes in radiology protocols can rapidly degrade the performance of static models. A continually adaptive FM could incrementally align its internal representations with new data, mitigating the need for repeated full retraining and reducing technical cost. Smith et al.~\cite{ref149_adaptive_replay} proposed an adaptive memory replay framework that dynamically selects the most relevant past samples via a bandit-based sampling strategy, effectively mitigating forgetting by up to 10\% without additional computational overhead. By enabling such dynamic adaptation, future FMs can remain clinically relevant over years of deployment, improving robustness to domain shifts while building greater trust among healthcare practitioners.

\subsection{Federated and Privacy-Preserving Adaptation}
Another critical barrier lies in the centralized nature of training, which often conflicts with the privacy and governance structures of medical institutions. Aggregating sensitive patient data in a single location is increasingly infeasible due to strict regulatory frameworks and growing awareness of risks around breaches and misuse. Federated learning (FL) offers an attractive alternative by enabling decentralized training across multiple sites without moving data. However, conventional FL often underperforms in medical imaging due to extreme data heterogeneity, scanner variation, and inconsistent acquisition protocols \cite{ref10_fm,ref150_pfl}. Future work must advance federated FM adaptation that is both domain-aware and communication-efficient, while integrating privacy-preserving safeguards such as secure aggregation and differential privacy. Recent innovations, such as pFLSynth \cite{ref150_pfl} have advanced FL by incorporating personalization blocks, partial model aggregation, and site-specific modulation layers, thereby improving cross-site performance in multi-contrast MRI synthesis. Building on such ideas, federated adaptation of lightweight modules (e.g., LoRA adapters or prompts) offers a scalable path forward, reducing communication costs and enabling site-specific customization. Coupled with federated continual learning, these approaches can ensure that FMs evolve collaboratively across institutions, achieving global robustness without compromising privacy, thereby enabling clinically trustworthy and widely deployable systems.

\subsection{Hybrid SSL for Data-Efficient Adaptation}
The scarcity and cost of high-quality annotations remain a central challenge in adapting FMs to medical imaging. While SSL has emerged as a promising solution, most existing approaches rely on single pretext tasks such as contrastive learning or masked image modeling. This narrow reliance fails to exploit the complementary strengths of different SSL paradigms. Recently, hybrid SSL strategies have emerged that combine multiple objectives, including MIM, contrastive learning, and cross-modal alignment. For example, integrating MIM (e.g., MAE-style reconstruction) with contrastive learning enables the model to jointly capture global semantic features and fine-grained spatial details, improving sample efficiency in downstream tasks. In multimodal settings, hybrid SSL methods can also learn joint representations across image–text modalities by aligning medical images with their associated radiology reports, even in the absence of explicit supervision \cite{ref115_cmae}. This is particularly advantageous in radiology, where large volumes of weakly annotated image–report pairs are available. Future research must focus on domain-adaptive hybrid SSL frameworks that dynamically balance multiple objectives, incorporate anatomical priors, and remain robust to noisy labels. Such approaches will be pivotal in building label-efficient FMs capable of generalizing across diverse modalities and tasks, reducing annotation costs while maintaining clinical fidelity.

\subsection{Data-Centric Adaptation with Synthetic and Human-in-the-Loop Pipelines}
Although advances in architectures and compute have been rapid, the progress of FM adaptation in medical imaging is increasingly constrained by the inadequacy of medical data in terms of scale, balance, or representation. Traditional methods such as manual labeling or simple augmentations are insufficient to address rare pathologies, underrepresented demographics, and scanner heterogeneity. Hence, synthetic data generation, particularly using diffusion and GAN-based models has emerged as a compelling strategy to expand datasets and mitigate these gaps. For example, RoentGen \cite{ref159_2024vision} demonstrated that domain-adapted latent diffusion models can boost downstream chest X-ray classification accuracy by 5\% and improve pneumothorax representation quality by 25\% when synthetic data is combined with real training samples. SynthRAD2023 \cite{ref160_3synthrad2023} addressed the parallel challenge of benchmarking scarcity by releasing a multi-center dataset and demonstrated that MRI-to-CT synthesis can maintain radiotherapy dose calculation errors within 1–2\% of planning CT, an error margin considered clinically acceptable for safety-critical workflows. However, synthetic data alone is insufficient, and future strategies must embed human-in-the-loop validation and active learning to ensure that generated cases are both pathologically plausible and clinically relevant. One promising direction is uncertainty-driven targeted generation, where models selectively synthesize data for underrepresented pathologies and iteratively refine these cases under expert review. Such data-centric approaches shift the focus from simply scaling model parameters to scaling the diversity, fairness, and quality of training data. By positioning data as the primary bottleneck, these human-guided synthetic pipelines can unlock the next wave of clinically robust and trustworthy FM adaptation.

\subsection{Benchmarking Generalization and Robustness in Real-World Clinical Scenarios} 

Despite rapid advances in FMs for MIA, a persistent gap remains between benchmark performance and real-world clinical utility. Most evaluations rely on curated datasets that are homogeneous, well-annotated, and limited in scope, failing to reflect the complexity, variability, and noise encountered in clinical practice. As a result, models that perform well on internal test sets often struggle with out-of-distribution data, rare conditions, or artifacts introduced by heterogeneous scanners. Moreover, current benchmarks also overemphasize accuracy metrics such as Dice or AUC, which are insufficient to capture critical dimensions of trustworthiness \cite{ref64_ft_distor_feat}. To bridge this gap, future benchmarking must move toward broader, task-diverse, and clinically validated protocols that reflect the realities of deployment. This includes evaluation across institutions, scanner types, and patient demographics, as well as stress-testing against missing data, label noise, and adversarial perturbations. Recent efforts such as MedMNIST \cite{ref151_medmnist}, MIMIC-CXR \cite{ref153_mimic}, and CheXpert \cite{ref152_chexpert} represent initial steps in this direction, but comprehensive and longitudinal benchmarks are required. In addition, evaluation must expand beyond accuracy to include calibration, fairness, uncertainty quantification, and failure mode analysis, which are essential for clinical adoption. Only through such rigorous and context-aware validation can we ensure that FMs not only achieve strong performance in the laboratory but also prove generalizable, reliable, and safe in clinical settings.

%% file: 07_conclusion.tex
\section{Conclusion}\label{sec7:conclusion}
FMs represent a transformative shift in MIA, offering unprecedented flexibility, scalability, and generalizability across a wide range of clinical tasks. This review has critically examined the architectural evolution, adaptation strategies, and deployment considerations that underpin the successful integration of FMs into medical workflows. From early transformer-based architectures to multimodal and self-supervised learning paradigms, the field has made significant strides in developing models that can leverage large-scale, heterogeneous data to deliver robust visual representations. However, persistent limitations continue to hinder real-world impact. Many current adaptations remain static, data-hungry, and insufficiently validated under clinical conditions. To address these challenges, we identified five converging trends that we argue will shape the next generation of FM adaptation: (1) continual learning for dynamic deployment, (2) federated and privacy-preserving adaptation, (3) hybrid self-supervised learning for data efficiency, (4) data-centric adaptation with synthetic and human-in-the-loop pipelines, (5) robust benchmarking aligned with real-world clinical variability. Together, these trends reflect a necessary shift from narrowly tuned, benchmark-driven models to adaptive, trustworthy, and clinically integrated systems. Moving forward, progress will require a holistic perspective that combines architectural innovation with pragmatic deployment strategies emphasizing data privacy, interpretability, and robustness across institutions and populations. Bridging this gap will demand interdisciplinary collaboration, standardized evaluation protocols, and continual feedback from real-world deployments, ultimately paving the way for clinically reliable and widely deployable FMs.